\pdfoutput=1
\documentclass[11pt]{article}

\usepackage[final]{acl}

\usepackage{times}
\usepackage{latexsym}

\usepackage[T1]{fontenc}
\usepackage[utf8]{inputenc}

\usepackage{microtype}

\usepackage{inconsolata}

\usepackage{colortbl}
\usepackage{graphicx}
\usepackage{xcolor}
\usepackage{url}
\usepackage{booktabs}
\usepackage{xspace}
\usepackage{multirow}
\usepackage{array}
\usepackage{tabularx}
\usepackage{pifont}
\usepackage{subcaption}

\newcommand{\data}{\texttt{TED2025}\xspace}

\title{From Unaligned to Aligned: Scaling Multilingual LLMs with Multi-Way Parallel Corpora}

\author{Yingli Shen$^{1}$\thanks{Equal contribution.}, Wen Lai$^{2,3*}$, Shuo Wang$^{1}$, Ge Gao$^{4}$ \\ 
\textbf {Kangyang Luo$^{1}$, Alexander Fraser$^{2,3}$, Maosong Sun$^{1,5}$\thanks{Corresponding author.} }\\[0.7em]
        \textsuperscript{1} Department of Computer Science and Technology, Tsinghua University \\
	\textsuperscript{2} Technical University of Munich \quad
        \textsuperscript{3} Munich Center for Machine Learning \\
        \textsuperscript{4} Minzu University of China \quad
        \textsuperscript{5} Institute for AI, Tsinghua University \\
\texttt{syl@mail.tsinghua.edu.cn, wen.lai@tum.de}
}

\begin{document}
\maketitle
\begin{abstract}
Continued pretraining and instruction tuning on large-scale multilingual data have proven to be effective in scaling large language models (LLMs) to low-resource languages.
However, the unaligned nature of such data limits its ability to effectively capture cross-lingual semantics.
In contrast, \emph{multi-way parallel data}, where identical content is aligned across multiple languages, provides stronger cross-lingual consistency and offers greater potential for improving multilingual performance.
In this paper, we introduce a large-scale, high-quality multi-way parallel corpus, \data, based on TED Talks.
The corpus spans 113 languages, with up to 50 languages aligned in parallel, ensuring extensive multilingual coverage.
Using this dataset, we investigate best practices for leveraging multi-way parallel data to enhance LLMs, including strategies for continued pretraining, instruction tuning, and the analysis of key influencing factors.
Experiments on six multilingual benchmarks show that models trained on multi-way parallel data consistently outperform those trained on unaligned multilingual data.\footnote{\url{https://github.com/yl-shen/multi-way-llm}}
\end{abstract}

\section{Introduction}
\label{sec:intro}
Large language models (LLMs) have demonstrated remarkable performance on various tasks in high-resource languages~\citep{huang2024survey,qin2024multilingual}. However, their performance still lags behind in low-resource languages~\citep{huang-etal-2023-languages,lai-etal-2023-chatgpt}.
To bridge this gap, recent efforts have focused on continued pretraining~\citep{ji2024emma,groeneveld-etal-2024-olmo} and instruction tuning~\citep{lai-etal-2024-llms,ustun-etal-2024-aya}, utilizing large-scale unaligned multilingual text.
However, these methods do not take full advantage of the explicit many-to-many alignments present in \emph{multi-way parallel corpora}, which have been shown to improve cross-lingual representations in multilingual NLP~\citep{qi-etal-2018-pre,freitag-firat-2020-complete,xu-etal-2022-eag,wu-etal-2024-far}.
More recently, \citet{mu-etal-2024-revealing} demonstrated that multi-way parallel inputs can also enhance in-context learning~\citep{dong-etal-2024-survey}.
However, scaling multilingual LLMs using multi-way parallel data remains underexplored.

%% multi-way dataset comparison
\definecolor{rowgray}{RGB}{248,248,248}    % 
\definecolor{myblue}{RGB}{224, 240, 255}   % 

\begin{table*}[t]
\resizebox{\textwidth}{!}{
\rowcolors{2}{}{rowgray}  
\arrayrulecolor{black!80}  
\begin{tabular}{lcccccccc}
\toprule
\rowcolor{white} 
             \textbf{Dataset} & \textbf{Source}      & \textbf{\#Langs} & \textbf{\#Domains} & \textbf{Max Parallelism} & \textbf{Type}       & \textbf{Collection Method} & \textbf{Date Range} & \textbf{Open-source?} \\
\midrule
Bible~\cite{christodouloupoulos2015massively}      & Bible       & 100     & 1         & 55              & Training   & Human Translator    & 2015       & \textcolor{green!50!black}{\ding{51}} \\
UN Corpus~\cite{ziemski-etal-2016-united}          & UN          & 6       & 1         & 6               & Training   & Human Translator    & 2016       & \textcolor{green!50!black}{\ding{51}} \\
MMCR4NLP~\cite{dabre2017mmcr4nlp}                  & Mix        & 59      & 5         & 13              & Training   & Mix Collection      & 2017       & \textcolor{red}{\ding{55}} \\
GCP Corpus~\cite{imamura-sumita-2018-multilingual} & Speech      & 10      & 4         & 10              & Training   & Machine Translation & 2018       & \textcolor{red}{\ding{55}} \\
FLORES-101~\cite{goyal-etal-2022-flores}           & Wiki        & 101     & 10        & 101             & Evaluation & Human Translator    & 2022       & \textcolor{green!50!black}{\ding{51}} \\
FLORES-200~\cite{costa2022no}                      & Wiki        & 204     & 10        & 204             & Evaluation & Human Translator    & 2022       & \textcolor{green!50!black}{\ding{51}} \\
BPCC~\cite{gala2023indictrans2}                    & Many        & 22      & 13        & 22              & Training   & Mix Collection      & 2023       & \textcolor{green!50!black}{\ding{51}} \\
XDailyDialog~\cite{liu-etal-2023-xdailydialog}     & DailyDialog & 4       & /         & 4               & Training   & Human Translator    & 2023       & \textcolor{green!50!black}{\ding{51}} \\
MWccMatrix~\cite{thompson-etal-2024-shocking}      & Common Crawl & 90      & /         & /               & Training   & Crawl               & 2024       & \textcolor{green!50!black}{\ding{51}} \\
TED2018~\cite{qi-etal-2018-pre}                    & TED         & 58      & /         & 58              & Training   & Human Translator    & 2018       & \textcolor{green!50!black}{\ding{51}} \\
TED2020~\cite{reimers-gurevych-2020-making}        & TED         & 108     & /         & /               & Training   & Human Translator    & 2020       & \textcolor{green!50!black}{\ding{51}} \\
\rowcolor{myblue}
\midrule
\textbf{Ours (\data)}                                         & TED         & 113     & 352       & 50              & Training   & Human Translator    & 2025       & \textcolor{green!50!black}{\ding{51}} \\
\bottomrule
\end{tabular}}
\vspace{-0.5em}
\caption{\label{tab:data_compare}
Comparison of existing multi-way parallel corpora and our constructed \data, highlighting key attributes such as the data source, the number of languages~(\#Langs), the number of domains~(\#Domains), the maximum parallelism supported, the type of data (training or evaluation), the collection method, and whether the corpus is open-source.
}
\vspace{-1em}
\end{table*}
Existing multi-way parallel datasets typically cover only a limited number of languages, domains and levels of parallelism (see Table~\ref{tab:data_compare}).
In contrast, TED Translators\footnote{\url{https://www.ted.com/participate/translate}}, a global community translating TED talk transcripts into over 100 languages, provides consistently high-quality, human-verified translations and serves as an ideal source for a large-scale multi-way parallel corpus.
However, the largest TED-based datasets~\citep{qi-etal-2018-pre,reimers-gurevych-2020-making} have not been updated since 2020, limiting their utility for LLM training and potentially exacerbating hallucinations~\citep{ji2023survey}.

To address these limitations, we introduce \data, a large-scale, high-quality multi-way parallel corpus derived from the latest TED talks.
\data covers 113 languages, 352 domain labels, and supports up to 50-way parallelism.
Compared to existing resources, it offers more frequent updates and significantly broader coverage across both languages and domains, thereby strengthening the data foundation for multilingual LLM training.

Utilizing \data, we investigate three key research questions: \par
\noindent \textbf{RQ1:} How does fine-tuning on multi-way parallel data compare to training on unaligned multilingual text in terms of zero-shot cross-lingual transfer and representation alignment? \par
\noindent \textbf{RQ2:} Which strategies for selecting parallelism in multi-way parallel data (e.g., degree of parallelism and language subsets) lead to the greatest improvements in multilingual LLM performance? \par
\noindent \textbf{RQ3:} Which instruction-tuning objectives can most effectively leverage the advantages of multi-way parallel data?

We perform a comprehensive evaluation across six multilingual benchmarks to assess the benefits of using multi-way parallel data for scaling multilingual LLMs.
Our results reveal that, at an equal data scale, fine-tuning on multi-way parallel data consistently outperforms training on unaligned multilingual text for both low-resource and high-resource languages (Section~\ref{sec:rq1}).
Additionally, we identify the most effective configurations of parallelism (Section~\ref{sec:rq2}).
Furthermore, we investigate how different instruction-tuning objectives impact LLM performance and cross-domain robustness (Section~\ref{sec:rq3}).

In summary, our contributions are as follows:
\textbf{(i)} We construct \data, a 50-way parallel corpus derived from recent TED talk transcripts, covering 113 languages and 352 domains.
\textbf{(ii)} To the best of our knowledge, this is the first work to leverage multi-way parallel data for scaling multilingual LLMs. We present a systematic comparison of multilingual LLM fine-tuning using multi-way versus unaligned data, analyzing their effects on zero-shot transfer and cross-lingual representation alignment.
\textbf{(iii)} We explore instruction-tuning objectives specifically designed for multi-way parallel data and provide practical recommendations for optimizing multilingual LLM performance.
\section{Related Work}
\label{sec:related_work}

\paragraph{Multi-Way Parallel Corpora.}
Datasets containing the same content across multiple languages (typically more than two) are known as \emph{multi-way parallel corpora}.
These corpora have demonstrated substantial benefits for machine translation~\citep{freitag-firat-2020-complete,xu-etal-2022-eag,wu-etal-2024-far} and cross-lingual representation alignment~\citep{tran2020cross}.
Existing methods for constructing such corpora include mining comparable texts~\citep{resnik1999bible}, aligning independently collected monolingual corpora via translation pivots~\citep{thompson-etal-2024-shocking}, extracting multi-way subsets from large bilingual collections~\citep{ramesh-etal-2022-samanantar}, and harvesting multilingual web crawls~\citep{resnik-smith-2003-web,qi-etal-2018-pre}.
However, many of these resources are limited in terms of language and domain coverage.
In contrast, we construct a multi-way parallel corpus derived from recent TED Talk transcripts, offering a broader and more diverse set of languages and domains.

\paragraph{Scaling Multilingual LLMs.}
The multilingual capabilities of LLMs have been significantly enhanced through two complementary strategies: continued pretraining on diverse multilingual corpora and multilingual instruction tuning.
Continued pretraining on unaligned multilingual data has improved both in-language fluency and cross-lingual transfer~\citep{ji2024emma,groeneveld-etal-2024-olmo}, while instruction tuning with human-curated multilingual prompts has boosted task performance across a wide range of languages~\citep{lai-etal-2024-llms,ustun-etal-2024-aya}.
More recently, \citet{mu-etal-2024-revealing} demonstrated that incorporating multi-way parallel examples into in-context prompts leads to further gains in zero-shot transfer.
Building on these insights, we systematically fine-tune multilingual LLMs on large-scale multi-way parallel data and quantify their impact compared to conventional unaligned approaches.
\section{Experimental Setup}
\label{sec:exp}
%% data statistics: parallelism
\begin{figure*}[t]
    \centering
    \includegraphics[width=\textwidth]{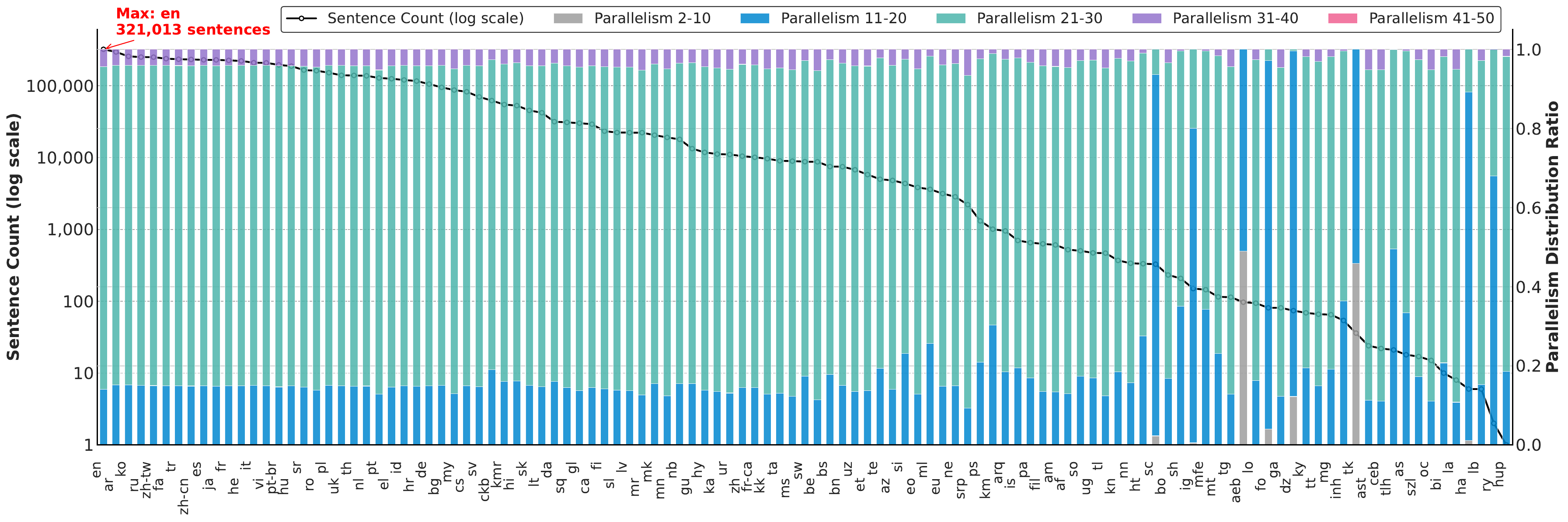}
    \caption{\label{fig:parallelism_sta}
    Distribution of sentence counts (line chart, left y-axis, log scale) and parallelism spans (bar chart, right y-axis, ratio) across languages (x-axis) in the \data{} corpus. The parallelism spans, with a notable concentration between 21 and 30 languages, and high range even for low-resource languages.}
    \vspace{-1em}
\end{figure*}

\begin{figure}[t]
    \centering
    \includegraphics[width=\columnwidth]{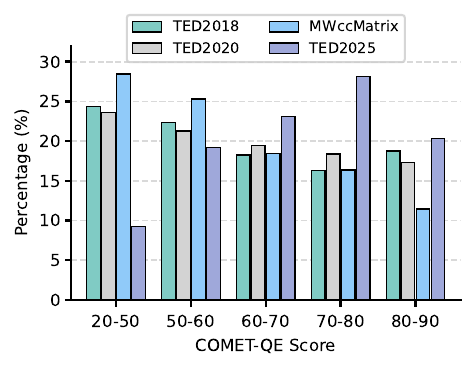}
    \caption{Comparison of translation quality between \data{} and existing multi-way datasets, including TED2018~\cite{qi-etal-2018-pre}, TED2020~\cite{reimers-gurevych-2020-making}, MWccMatrix~\cite{thompson-etal-2024-shocking}, using COMET-QE score.
}
    \label{fig:qe_sta}
\end{figure}

\paragraph{\data.}
We introduce \data, a new multi-way parallel corpus derived from the latest TED Talk transcripts.
It encompasses 113 languages with up to 50-way parallelism, making it one of the largest and most diverse resources for multilingual fine-tuning. 
Figure~\ref{fig:parallelism_sta} illustrates the total number of sentences and the distribution of parallelism spans across languages in \data. 
Figure~\ref{fig:qe_sta} compares translation quality, as measured by COMET-QE~\citep{rei-etal-2020-comet}, across \data (which includes 4,765 language pairs\footnote{For language pairs with more than 10,000 sentence pairs, we randomly sample 10,000.}) and other existing multi-way datasets: TED2018~\citep{qi-etal-2018-pre}, TED2020~\citep{reimers-gurevych-2020-making}, and MWccMatrix~\citep{thompson-etal-2024-shocking}.
Additional dataset statistics and construction details are provided in Appendix~\ref{app:dataset}.

We observe that:
(1) the most common parallelism span in \data ranges from 21 to 30 languages. Notably, many low-resource languages also achieve high degrees of parallelism, providing a solid foundation for multilingual research.
(2) \data contains significantly more high-quality translations (with a COMET-QE score greater than 60) compared to previous multi-way corpora.
Unlike TED2020, which segments English based on punctuation, or MWccMatrix, which relies on LASER score~\citep{artetxe-schwenk-2019-massively}, we use human-provided timestamps to generate cleaner and more reliable sentence alignments.

%% Figure: QE quality

\paragraph{Training.}
To isolate the effects of multi-way parallel data, we conduct both continued pretraining~\citep{parmar2024reuse} and instruction tuning~\citep{zhang2023instruction} on \data.
% \footnote{For RQ1 and RQ2, we focus on continued pretraining to avoid interference from instruction tuning. For RQ3, we instead focus on instruction tuning to assess its specific impact.}.
We experiment with two model families and sizes: LLaMA-3.1-8B / LLaMA-3.1-8B-Instruct and Qwen-2.5-14B / Qwen-2.5-14B-Instruct (available on Hugging Face\footnote{\url{https://huggingface.co}}).
To make fine-tuning feasible, we employ Low-Rank Adaptation (LoRA)~\citep{hu2022lora} instead of performing full parameter updates.
Full hyperparameter settings and training configurations are provided in Appendix~\ref{app:exp}.

%% Table
\begin{table}[t]
\resizebox{\columnwidth}{!}{
\begin{tabular}{lccc}
\toprule
\textbf{Benchmark}   & \textbf{Task}                  & \textbf{\#Langs} & \textbf{Metric}     \\
\midrule
\textbf{MMMLU}       & Understanding         & 14      & Acc        \\
\textbf{XCOPA}       & Reasoning             & 11      & Acc        \\
\textbf{FLORES-101}  & Generation            & 101     & BLEU/COMET \\
\textbf{FLORES-200}  & Generation            & 204     & BLEU/COMET \\
\textbf{xIFEval} & Instruction Following & 17      & Acc   \\
\textbf{SIB-200} & Text Classification & 204 & Acc  \\
\bottomrule
\end{tabular}}
\caption{
\label{tab:eval_metric}
Overview of evaluation benchmarks, including task types, the number of languages (\#Langs) involved, and the metrics used for assessment.
}
\vspace{-1em}
\end{table}

\paragraph{Evaluation and Metrics.}
We evaluate our models on five widely adopted multilingual benchmarks in a zero-shot setting, covering a range of tasks: understanding (MMMLU), reasoning (XCOPA; \citealp{ponti-etal-2020-xcopa}), generation (FLORES-101; \citealp{goyal-etal-2022-flores} and FLORES-200; \citealp{costa2022no}), instruction following (xIFEval; \citealp{huang2025benchmax}), and text classification (SIB-200; \citealp{adelani-etal-2024-sib}).
Table~\ref{tab:eval_metric} summarizes these benchmarks along with their associated evaluation metrics.
Additionally, we categorize the languages in each benchmark as low-resource or high-resource based on the classification in~\citet{costa2022no}.
\section{Effectiveness of Multi-Way Corpora}
\label{sec:rq1}
We investigate the impact of training on multi-way parallel data on the performance of a multilingual LLM across three key dimensions:
downstream performance on multilingual benchmarks (Section~\ref{sec:rq1_downstream}), zero-shot cross-lingual transfer to unseen languages (Section~\ref{sec:rq1_cross_lingual}) and cross-lingual alignment of representations within the model’s internal embeddings (Section~\ref{sec:rq1_alignment}).

For fairness, we fix the total continued pretraining data at 5 million tokens (5M) and evaluate two LLM backbones (LLaMA-3-8B and Qwen-2.5-14B) under three conditions:
(1) \textbf{Multi-Way}: Pretraining on our multi-way parallel corpus (\data);
(2) \textbf{Unaligned}: Pretraining on an unaligned multilingual corpus\footnote{Unaligned data is sourced from DCAD-2000~\cite{shen2025dcad} rather than \data, due to considerations related to scalability and domain coverage.} (DCAD-2000; \citealp{shen2025dcad});
(3) \textbf{Baseline}: The original pretrained checkpoint without additional data.

%%% Table: RQ1
\definecolor{rowgray}{RGB}{240,240,240}    
\definecolor{highlight}{RGB}{224, 240, 255}  % 定义浅蓝色

\begin{table*}[t]
\centering
\begin{subtable}{\textwidth}
\centering
\resizebox{\textwidth}{!}{
\begin{tabular}{l|cc|cc|cccc|cccc|cc}
\toprule
& \multicolumn{2}{c}{\multirow{2}{*}{\textbf{MMMLU}}} & \multicolumn{2}{c}{\multirow{2}{*}{\textbf{XCOPA}}} & \multicolumn{4}{c}{\textbf{FLORES-101 (Eng-X)}} & \multicolumn{4}{c}{\textbf{FLORES-101 (X-Eng)}} & \multicolumn{2}{c}{\multirow{2}{*}{\textbf{xIFEval}}} \\
\cmidrule(lr){6-9}\cmidrule(lr){10-13}
& \multicolumn{2}{l}{} & \multicolumn{2}{l}{} & \multicolumn{2}{c}{\textbf{BLEU}} & \multicolumn{2}{c}{\textbf{COMET}} & \multicolumn{2}{c}{\textbf{BLEU}} & \multicolumn{2}{c}{\textbf{COMET}} & \multicolumn{2}{l}{} \\
\cmidrule(lr){6-7}\cmidrule(lr){8-9}\cmidrule(lr){10-11}\cmidrule(lr){12-13}
& low & high & low & high & low & high & low & high & low & high & low & high & low & high \\
\midrule
\rowcolor{rowgray}
Baseline  & 18.27 & 33.72 & 23.46 & 34.29 & 6.03 & 11.67 & 57.15 & 61.03 & 13.37 & 22.49 & 75.24 & 82.32 & 17.14 & 24.43 \\

Unaligned & 19.64 & 36.26 & 24.62 & 34.76 & 6.12 & 11.78 & 57.51 & 62.11 & 13.84 & 22.74 & 75.82 & 82.58 & 17.28 & 24.44 \\
\rowcolor{highlight}
\textbf{Multi-Way} & \textbf{22.48} & \textbf{41.38} & \textbf{27.58} & \textbf{57.22} & \textbf{6.32} & \textbf{12.08} & \textbf{58.06} & \textbf{67.44} & \textbf{14.45} & \textbf{25.03} & \textbf{76.25} & \textbf{86.43} & \textbf{18.79} & \textbf{27.41} \\
\bottomrule
\end{tabular}}
\caption{LLaMA-3-8B}
\label{tab:rq1_downstream_a}
\end{subtable}

\vspace{0.15cm}

\begin{subtable}{\textwidth}
\centering
\resizebox{\textwidth}{!}{
\begin{tabular}{l|cc|cc|cccc|cccc|cc}
\toprule
& \multicolumn{2}{c}{\multirow{2}{*}{\textbf{MMMLU}}} & \multicolumn{2}{c}{\multirow{2}{*}{\textbf{XCOPA}}} & \multicolumn{4}{c}{\textbf{FLORES-101 (Eng-X)}} & \multicolumn{4}{c}{\textbf{FLORES-101 (X-Eng)}} & \multicolumn{2}{c}{\multirow{2}{*}{\textbf{xIFEval}}} \\
\cmidrule(lr){6-9}\cmidrule(lr){10-13}
& \multicolumn{2}{l}{} & \multicolumn{2}{l}{} & \multicolumn{2}{c}{\textbf{BLEU}} & \multicolumn{2}{c}{\textbf{COMET}} & \multicolumn{2}{c}{\textbf{BLEU}} & \multicolumn{2}{c}{\textbf{COMET}} & \multicolumn{2}{l}{} \\
\cmidrule(lr){6-7}\cmidrule(lr){8-9}\cmidrule(lr){10-11}\cmidrule(lr){12-13}
& low & high & low & high & low & high & low & high & low & high & low & high & low & high \\
\midrule
\rowcolor{rowgray}
Baseline  & 35.24 & 49.55 & 62.25 & 72.00 & 7.45 & 11.05 & 57.22 & 67.16 & 16.54 & 20.24 & 67.23 & 74.29 & 27.63 & 32.40 \\

Unaligned & 35.61 & 51.32 & 62.59 & 74.06 & 7.62 & 11.60 & 57.85 & 70.85 & 16.86 & 21.02 & 67.61 & 75.97 & 27.92 & 35.54 \\
\rowcolor{highlight}
\textbf{Multi-Way} & \textbf{36.64} & \textbf{55.81} & \textbf{63.24} & \textbf{79.52} & \textbf{8.07} & \textbf{13.11} & \textbf{58.94} & \textbf{80.56} & \textbf{17.36} & \textbf{23.26} & \textbf{68.59} & \textbf{81.33} & \textbf{28.64} & \textbf{40.95} \\
\bottomrule
\end{tabular}}
\caption{Qwen-2.5-14B}
\label{tab:rq1_downstream_b}
\end{subtable}
\caption{Performance~(\%) comparison of different models across multilingual benchmarks. The \textbf{Multi-Way} approach with blue background demonstrates consistent superiority in both low-resource (left columns) and high-resource (right columns) scenarios.}
\label{tab:rq1_downstream}
\end{table*}
%%%

\subsection{Downstream Performance}
\label{sec:rq1_downstream}
We evaluate all variants in a zero-shot setting across four benchmarks—MMMLU, XCOPA, FLORES-101, and xIFEval—that cover understanding, reasoning, generation, and instruction following.
The full results are reported in Table~\ref{tab:rq1_downstream}.
Our findings show that, across all tasks, \textit{Multi-Way} consistently outperforms both \textit{Baseline} and \textit{Unaligned} across both low- and high-resource languages.
For example, on MMMLU, Multi-Way achieves accuracies of 22.48/41.38 in low/high-resource languages, compared to 18.27/33.72 for the \textit{Baseline} and 19.64/36.26 for \textit{Unaligned}.
Similar improvements are observed on FLORES-101 and xIFEval.
These results demonstrate that multi-way alignment provides stronger cross-lingual supervision, thereby enhancing both discriminative and generative capabilities.
Although our main experiments focus on English–X translation directions, we also evaluate the trained models on non-English-centric translation pairs. The results, which show that models trained on aligned data significantly outperform those trained on unaligned data even for non-English-centric tasks, are provided in Appendix~\ref{app:non_eng}.

%%% Figure RQ1: cross-lingual transfer
\begin{figure}[ht]
    \centering
    \includegraphics[width=\linewidth]{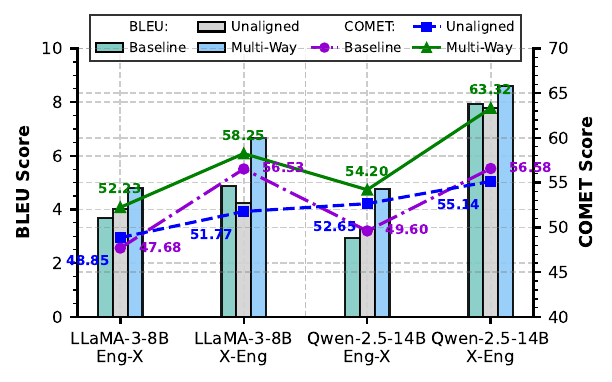}
    \caption{\label{fig:cross-transfer}
    Cross-lingual transfer performance comparison between Baseline, Unaligned and Multi-Way pretraining on the FLORES-200 benchmark with BLEU (bar chart, left y-axis) and COMET (line chart, right y-axis) for LLaMA-3-8B and Qwen-2.5-14B models.}
    \vspace{-1em}
\end{figure}

\subsection{Zero-Shot Cross-Lingual Transfer}
\label{sec:rq1_cross_lingual}
To assess generalization to unseen languages~\cite{lai-etal-2022-m4,zhao-etal-2025-adamergex}, we evaluate on FLORES‑200.
We exclude all languages in the evaluation subset from training and assess the English$\leftrightarrow$X translation quality.
As shown in Figure~\ref{fig:cross-transfer}, the \textit{Multi‑Way} model significantly outperforms both \textit{Baseline} and \textit{Unaligned} in both translation directions.
This highlights that explicit multi-way supervision promotes language-agnostic representations, enabling robust zero-shot transfer.
We further explore this hypothesis in Section~\ref{sec:rq1_alignment}, where we analyze differences in cross-lingual representation alignment across models.

%% Table RQ1: alignment

\definecolor{myred}{RGB}{200,0,0}
\definecolor{myblue}{RGB}{0,0,150}

\begin{table*}[ht]
\centering
\small
\begin{tabular}{llccc|ccc}
\toprule
                           &      & \multicolumn{3}{c}{\textbf{LLaMA-3-8B}}     & \multicolumn{3}{c}{\textbf{Qwen-2.5-14B}}   \\
\cmidrule(lr){3-5}\cmidrule(lr){6-8}
                           &      & Baseline & Unaligned & Multi-Way & Baseline & Unaligned & Multi-Way \\
\midrule
\multicolumn{2}{l}{\textbf{Cosine ($\uparrow$)}}        
& 0.27
& 0.27\raisebox{-3pt}{\textcolor{myblue}{\scriptsize{-0.00}}} 
& \textbf{0.30}\raisebox{-3pt}{\textcolor{myred}{\scriptsize{+0.03}}} 
& 0.29
& 0.27\raisebox{-3pt}{\textcolor{myblue}{\scriptsize{-0.02}}} 
& \textbf{0.32}\raisebox{-3pt}{\textcolor{myred}{\scriptsize{+0.03}}} \\

\midrule
\multicolumn{2}{l}{\textbf{CKA ($\uparrow$)}}    
& 0.54
& 0.54\raisebox{-3pt}{\textcolor{myblue}{\scriptsize{-0.00}}} 
& \textbf{0.60}\raisebox{-3pt}{\textcolor{myred}{\scriptsize{+0.06}}} 
& 0.56
& 0.57\raisebox{-3pt}{\textcolor{myred}{\scriptsize{+0.01}}} 
& \textbf{0.63}\raisebox{-3pt}{\textcolor{myred}{\scriptsize{+0.07}}} \\

\midrule
\multirow{3}{*}{\textbf{Retrieval ($\uparrow$)}} 
& P@1  
& 0.09
& 0.07\raisebox{-3pt}{\textcolor{myblue}{\scriptsize{-0.02}}} 
& \textbf{0.13}\raisebox{-3pt}{\textcolor{myred}{\scriptsize{+0.04}}} 
& 0.12
& 0.14\raisebox{-3pt}{\textcolor{myred}{\scriptsize{+0.02}}} 
& \textbf{0.19}\raisebox{-3pt}{\textcolor{myred}{\scriptsize{+0.07}}} \\

                           & P@5  
& 0.24
& 0.23\raisebox{-3pt}{\textcolor{myblue}{\scriptsize{-0.01}}} 
& \textbf{0.27}\raisebox{-3pt}{\textcolor{myred}{\scriptsize{+0.03}}} 
& 0.26
& 0.28\raisebox{-3pt}{\textcolor{myred}{\scriptsize{+0.02}}} 
& \textbf{0.33}\raisebox{-3pt}{\textcolor{myred}{\scriptsize{+0.07}}} \\

                           & P@10 
& 0.35
& 0.33\raisebox{-3pt}{\textcolor{myblue}{\scriptsize{-0.02}}} 
& \textbf{0.39}\raisebox{-3pt}{\textcolor{myred}{\scriptsize{+0.04}}} 
& 0.36
& 0.38\raisebox{-3pt}{\textcolor{myred}{\scriptsize{+0.02}}} 
& \textbf{0.42}\raisebox{-3pt}{\textcolor{myred}{\scriptsize{+0.06}}} \\

\midrule
\multicolumn{2}{l}{\textbf{SVCCA ($\uparrow$)}}         
& 0.55
& 0.55\raisebox{-3pt}{\textcolor{myblue}{\scriptsize{-0.00}}} 
& \textbf{0.61}\raisebox{-3pt}{\textcolor{myred}{\scriptsize{+0.06}}} 
& 0.57
& 0.58\raisebox{-3pt}{\textcolor{myred}{\scriptsize{+0.01}}} 
& \textbf{0.63}\raisebox{-3pt}{\textcolor{myred}{\scriptsize{+0.06}}} \\
\bottomrule
\end{tabular}
\caption{\label{tab:rq1_alignment}
Cross-lingual representation alignment results. Improvement margins (colored red/blue) are shown relative to Baseline. \textbf{Bolded} Multi-Way results with \textcolor{myred}{red} annotations indicate consistent improvements, particularly in cross-lingual retrieval tasks (e.g., +0.07 P@1 improvement for Qwen-2.5-14B).
}
\vspace{-1em}
\end{table*}

\subsection{Cross-Lingual Representation Alignment}
\label{sec:rq1_alignment}
We further analyze the alignment of internal representations across models.
To be specific, for 32 randomly selected aligned languages (with 100 sentences each) from \data, we compute the following four metrics:
average \emph{cosine similarity} between parallel sentence embeddings, \emph{Centered Kernel Alignment} (CKA) between representation matrices~\citep{kornblith2019similarity}, Cross-lingual sentence retrieval accuracy at P@1, P@5, and P@10~\citep{conneau2017word} and SVCCA score~\citep{raghu2017svcca}.

\begin{figure*}[t]
    \centering
    \includegraphics[width=\linewidth]{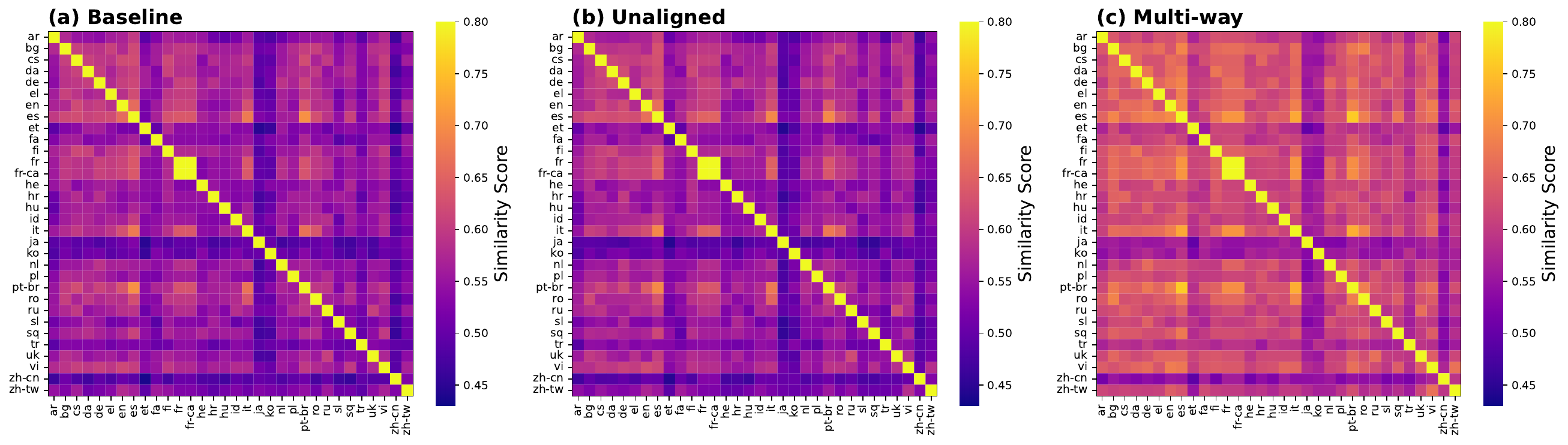}
    \caption{\label{fig:svcca_align}
    SVCCA alignment comparison between the Multi-Way, Unaligned and Baseline models across 32-way language pairs.}
    \vspace{-1em}
\end{figure*}

Table~\ref{tab:rq1_alignment} shows that \textit{Multi‑Way} outperforms the other models, yielding higher CKA and better retrieval accuracy.
Figure~\ref{fig:svcca_align} further corroborates these results: \textit{Multi‑Way} demonstrates denser SVCCA alignments, particularly for linguistically distant language pairs.
These metrics confirm that multi-way pretraining promotes a more coherent, language-agnostic embedding space, which drives the observed improvements in downstream performance and cross-lingual transfer.

\section{Impact Factors}
\label{sec:rq2}
In Section~\ref{sec:rq1}, we demonstrate that using multi-way parallel data can significantly enhance the multilingual capabilities of LLMs.
To better understand the factors driving this improvement, we analyze two key aspects, while keeping the total pretraining fixed at 5 million tokens:
(1) \textbf{Degree of parallelism}: the number of languages aligned in each training example (Section~\ref{sec:rq2_parallelism}).
(2) \textbf{English as a pivot}: the impact of including versus excluding English in multi-way groups (Section~\ref{sec:rq2_pivot}).
(3) \textbf{Language combinations}: the role of language family composition in training data (Section~\ref{sec:rq2_lang_family}).
(4) \textbf{Training data size}: how the amount of pretraining data affects model performance (Section~\ref{sec:rq2_data_size}).

%% Figure RQ1: SVCCA

\subsection{Degree of Parallelism}
\label{sec:rq2_parallelism}
\begin{figure}[t]
    \centering
    \includegraphics[width=\linewidth]{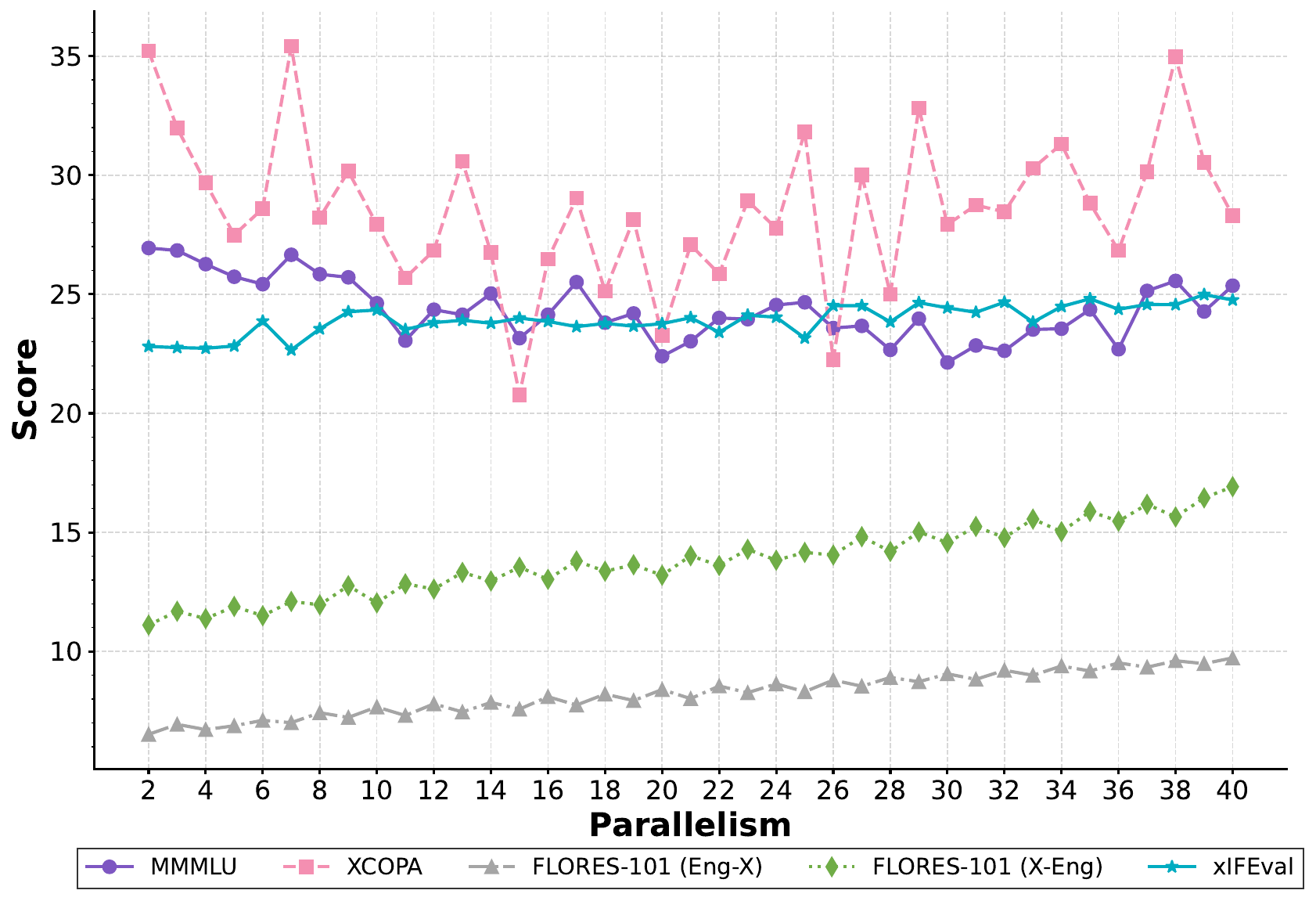}
    \caption{\label{fig:diff_parallelism}
    Performance (\%) of continued pretraining models on downstream tasks with varying degrees of parallelism.}
    \vspace{-1em}
\end{figure}

We construct datasets with parallelism levels ranging from 2 to 40 languages per example, sampled from the \data corpus, while always keeping 5M tokens.
Figure~\ref{fig:diff_parallelism} shows model performance across a range of tasks for each setting.
For bidirectional machine translation (FLORES-101, Eng$\rightarrow$X and X$\rightarrow$Eng), performance steadily improves with higher parallelism, suggesting that broader semantic alignment enhances cross-lingual generation and fluency.
In contrast, for non-generative tasks (reasoning and understanding), accuracy tends to peak at small parallelism (around 6–10 languages) before deteriorating.
We attribute this decline to two factors:
(1) Excessive linguistic diversity can obscure shared semantic patterns.
(2) With a fixed token budget, each language receives fewer tokens, limiting the ability to learn language-specific features.

%%% figure: different parallelism

\subsection{English as Pivot}
\label{sec:rq2_pivot}
English is widely used as a pivot language in multilingual MT and NLP~\cite{kim-etal-2019-pivot, mallinson-etal-2018-sentence,lai-etal-2023-mitigating}.
To explore its impact, we form five groups\footnote{We provide the configurations of each group in Table~\ref{tab:app_lang_comb} of Appendix~\ref{app:sec_lang_comb}}, each with both English-included and English-excluded variants across ten languages.
Table~\ref{tab:pivot} compares their performance across various tasks.

In tasks involving understanding and reasoning, groups that include English consistently outperform those in other language groups by an average of 2–4 percentage points.
This suggests that English, as a high-resource ``semantic anchor'', helps stabilize embedding alignment and facilitates transfer learning, particularly on complex tasks.
Interestingly, for machine translation (FLORES-101) and some instruction-following tasks (xIFEval), English-inclusive groups show slightly lower performance.
We attribute this to two primary factors:
(1) English occupies tokens that could otherwise be used to align non-English language pairs directly.
(2) The model may overly rely on English as an intermediary, which reduces its ability to directly transfer knowledge between non-English languages.
These findings highlight that English's role as a pivot language is task-dependent.
While it can enhance semantic coherence in understanding and reasoning tasks, it may hinder direct multilingual transfer in generative tasks. 
Consequently, the inclusion of English in multilingual pretraining should be carefully considered based on the target application.

%% figure pivot or not
% \input{figure/rq2_pivot}
\begin{table}[]
\resizebox{\columnwidth}{!}{
\begin{tabular}{ccccccc}
\toprule
                         &      & \textbf{MMMLU} & \textbf{XCOPA} & \begin{tabular}[c]{@{}l@{}}\textbf{FLORES}\\ \textbf{(Eng-X)}\end{tabular} & \begin{tabular}[c]{@{}l@{}}\textbf{FLORES}\\ \textbf{(X-Eng)}\end{tabular} & \textbf{xIFEval}  \\ 
\midrule
\multirow{2}{*}[\dimexpr-\baselineskip/2]{\centering \textbf{Group 1}} & with & \textbf{23.17} & \textbf{30.93} & 5.80               & 13.75              & 23.39   \\
                         & w/o  & 20.84 & 30.07 & \textbf{6.63}               & \textbf{14.56}              & \textbf{24.14}   \\
\midrule
\multirow{2}{*}[\dimexpr-\baselineskip/2]{\centering \textbf{Group 2}} & with & \textbf{22.19} & \textbf{35.33} & 6.74               & 13.88              & \textbf{22.19}   \\
                         & w/o  & 18.51 & 31.60 & \textbf{7.13}               & \textbf{14.18}              & \textbf{22.19}   \\
\midrule
\multirow{2}{*}[\dimexpr-\baselineskip/2]{\centering \textbf{Group 3}} & with & \textbf{26.47} & \textbf{36.73} & \textbf{6.40}               & 13.48              & 22.69   \\
                         & w/o  & 23.83 & 35.78 & 6.38               & \textbf{14.00}              & \textbf{22.91}   \\
\midrule
\multirow{2}{*}[\dimexpr-\baselineskip/2]{\centering \textbf{Group 4}} & with & \textbf{23.42} & \textbf{33.84} & 6.20               & 12.11              & 23.66   \\
                         & w/o  & 20.26 & 30.84 & \textbf{6.67}               & \textbf{14.76}              & \textbf{23.67}   \\
 \midrule
\multirow{2}{*}[\dimexpr-\baselineskip/2]{\centering \textbf{Group 5}} & with & \textbf{23.89} & \textbf{39.64} & 6.96               & 13.07              & 23.19   \\
                         & w/o  & 20.39 & 34.15 & \textbf{7.83}               & \textbf{14.79}              & \textbf{23.85}  \\
\bottomrule
\end{tabular}}
\caption{
\label{tab:pivot}
Performance~(\%) comparison of models with and without (w/o) English across five different language groupings.
}
\end{table}

%% diff objective table
\begin{table*}[!ht]
\centering
\begin{subtable}{\textwidth}
\centering
\resizebox{\textwidth}{!}{
\begin{tabular}{l|cc|cc|cccc|cccc|cc}
\toprule
& \multicolumn{2}{c}{\multirow{2}{*}{\textbf{MMMLU}}} & \multicolumn{2}{c}{\multirow{2}{*}{\textbf{XCOPA}}} & \multicolumn{4}{c}{\textbf{FLORES-101 (Eng-X)}}               & \multicolumn{4}{c}{\textbf{FLORES-101 (X-Eng)}}               & \multicolumn{2}{c}{\multirow{2}{*}{\textbf{xIFEval}}} \\
\cmidrule(lr){6-9}\cmidrule(lr){10-13}
& \multicolumn{2}{l}{}                       & \multicolumn{2}{l}{}                       & \multicolumn{2}{c}{\textbf{BLEU}} & \multicolumn{2}{c}{\textbf{COMET}} & \multicolumn{2}{c}{\textbf{BLEU}} & \multicolumn{2}{c}{\textbf{COMET}} & \multicolumn{2}{l}{}                         \\
\cmidrule(lr){6-7}\cmidrule(lr){8-9}\cmidrule(lr){10-11}\cmidrule(lr){12-13}
& low               & high                   & low               & high                   & low        & high        & low         & high        & low         & high       & low         & high        & low                & high                    \\
\midrule
Baseline              & 41.96                & 46.68               & 64.59                & 66.79               & 5.51        & 10.50      & 56.95       & 60.02       & 14.98       & 18.41      & 68.29       & 73.51       & 35.99                 & 38.56                \\
\midrule
MT                    & \textbf{45.28} & \textbf{51.06} & \textbf{68.17} & \textbf{69.38} & \textbf{11.26} & \textbf{13.25} & \textbf{63.03} & \textbf{65.61} & \textbf{22.25} & \textbf{21.60} & \textbf{70.76} & \textbf{75.41} & \textbf{41.72} & \textbf{44.52} \\
CLTS                  & 39.99                & 45.44               & 62.87                & 66.47               & 3.63        & 9.90       & 56.48       & 59.28       & 13.25       & 16.60      & 66.89       & 73.15       & 34.91                 & 38.38                \\
MTC                   & 40.68                & 44.49               & 64.19                & 65.16               & 5.02        & 9.08       & 56.04       & 57.84       & 14.38       & 18.16      & 67.86       & 73.12       & 34.03                 & 38.01                \\
CLP                   & 42.49                & 47.01               & 65.41                & 68.42               & 6.38        & 11.46      & 57.23       & 61.28       & 15.76       & 19.27      & 69.87       & 73.91       & 36.17                 & 40.07                \\
\midrule
MT + CLTS             & 42.23                & 47.94               & 65.00                & \textbf{68.77}               & 6.19        & 10.51      & \textbf{58.53}       & 60.33       & 16.83       & 18.98      & 68.70       & \textbf{74.91}       & 36.47                 & \textbf{40.30}                \\
MT + MTC              & 42.69                & 47.54               & 65.12                & 66.83               & 6.35        & 10.73      & 57.53       & 60.24       & 15.13       & 18.45      & \textbf{69.28}       & 74.00       & 36.10                 & 39.03                \\
MT + CLP              & 43.20                & \textbf{49.07}               & \textbf{67.29}                & 68.47               & \textbf{7.31}        & 10.51      & 58.40       & \textbf{62.64}       & \textbf{17.75}       & \textbf{20.77}      & 69.18       & 74.47       & \textbf{38.49}                 & 39.67                \\
CLTS + MTC            & 41.78                & 45.17               & 63.21                & 65.62               & 5.38        & 9.77       & 55.19       & 58.21       & 14.67       & 16.59      & 67.45       & 72.35       & 34.57                 & 36.63                \\
CLTS + CLP            & \textbf{42.82}                & 46.74               & 65.30                & 67.12               & 5.58        & \textbf{10.88}      & 57.84       & 60.41       & 15.41       & 19.33      & 68.84       & 73.58       & 36.99                 & 38.83                \\
MTC + CLP             & 42.62                & 46.70               & 65.16                & 67.56               & 6.48        & 10.84      & 57.74       & 60.98       & 15.29       & 19.22      & 68.76       & 73.77       & 36.87                 & 38.96                \\
\midrule
MT + CLTS + MTC       & 41.03                & 46.59               & 64.15                & 66.05               & 5.14        & 10.35      & 56.14       & 59.71       & 14.64       & 18.06      & 67.64       & 73.39       & 35.24                 & 38.14                \\
MT + CLTS + CLP       & \textbf{42.72}                & \textbf{47.67}               & 64.83                & 67.57               & \textbf{6.17}        & 10.94      & 57.78       & 60.11       & \textbf{15.96}       & \textbf{19.16}      & 68.67       & \textbf{74.03}       & \textbf{36.54}                 & 38.77                \\
MT + MTC + CLP        & 42.33                & 46.72               & \textbf{65.26}                & \textbf{67.58}               & 5.92        & \textbf{10.98}      & \textbf{57.93}       & \textbf{60.76}       & 15.38       & 18.81      & \textbf{68.98}       & 73.55       & 36.22                 & \textbf{39.15}                \\
CLTS + MTC + CLP      & 41.56                & 46.65               & 64.16                & 66.45               & 4.67        & 10.38      & 56.20       & 59.63       & 14.95       & 17.62      & 68.00       & 73.03       & 35.27                 & 38.21                \\
\midrule
MT + CLTS + MTC + CLP & 43.07                & 47.67               & 66.64                & 67.38               & 7.59        & 12.42      & 57.75       & 60.79       & 17.62       & 19.62      & 71.13       & 75.24       & 36.95                 & 40.52             \\
\bottomrule
\end{tabular}}
\caption{LLaMA-3-8B-Instruct}
\label{tab:rq3_object_a}
\end{subtable}

\vspace{0.15cm}

\begin{subtable}{\textwidth}
\centering
\resizebox{\textwidth}{!}{
\begin{tabular}{l|cc|cc|cccc|cccc|cc}
\toprule
& \multicolumn{2}{c}{\multirow{2}{*}{\textbf{MMMLU}}} & \multicolumn{2}{c}{\multirow{2}{*}{\textbf{XCOPA}}} & \multicolumn{4}{c}{\textbf{FLORES-101 (Eng-X)}} & \multicolumn{4}{c}{\textbf{FLORES-101 (X-Eng)}} & \multicolumn{2}{c}{\multirow{2}{*}{\textbf{xIFEval}}} \\
\cmidrule(lr){6-9}\cmidrule(lr){10-13}
& \multicolumn{2}{l}{} & \multicolumn{2}{l}{} & \multicolumn{2}{c}{\textbf{BLEU}} & \multicolumn{2}{c}{\textbf{COMET}} & \multicolumn{2}{c}{\textbf{BLEU}} & \multicolumn{2}{c}{\textbf{COMET}} & \multicolumn{2}{l}{} \\
\cmidrule(lr){6-7}\cmidrule(lr){8-9}\cmidrule(lr){10-11}\cmidrule(lr){12-13}
& low & high & low & high & low & high & low & high & low & high & low & high & low & high \\
\midrule
Baseline & 63.61 & 68.14 & 78.76 & 82.22 & 7.92 & 12.79 & 61.49 & 66.64 & 16.76 & 20.48 & 68.16 & 70.32 & 47.68 & 52.40 \\
\midrule
MT & \textbf{68.52} & \textbf{72.83} & \textbf{82.95} & \textbf{86.95} & \textbf{11.90} & \textbf{16.60} & \textbf{64.96} & \textbf{70.90} & \textbf{20.43} & \textbf{26.22} & \textbf{72.25} & \textbf{73.51} & \textbf{53.24} & \textbf{55.83} \\
CLTS & 61.28 & 67.61 & 76.77 & 79.74 & 7.20 & 10.70 & 60.90 & 64.50 & 15.75 & 18.21 & 67.01 & 69.83 & 45.73 & 51.83 \\
MTC & 62.59 & 65.63 & 76.87 & 81.46 & 6.07 & 12.29 & 61.48 & 65.61 & 13.84 & 19.71 & 67.12 & 68.65 & 47.53 & 50.53 \\
CLP & 64.28 & 70.06 & 79.69 & 83.02 & 9.54 & 13.10 & 62.18 & 68.35 & 17.54 & 21.68 & 69.24 & 70.68 & 48.61 & 52.66 \\
\midrule
MT + CLTS & 63.93 & 69.53 & 78.94 & 82.75 & 8.73 & 13.91 & 61.87 & 67.09 & 18.20 & 21.50 & 68.45 & 71.90 & 48.16 & \textbf{54.14} \\
MT + MTC & \textbf{65.33} & 69.12 & 80.27 & 83.34 & 8.64 & \textbf{13.94} & 61.71 & 67.42 & 17.87 & \textbf{22.27} & 68.60 & \textbf{72.01} & \textbf{49.46} & 53.53 \\
MT + CLP & 64.95 & \textbf{70.51} & \textbf{80.68} & \textbf{83.52} & \textbf{10.30} & 13.65 & \textbf{62.37} & \textbf{69.32} & \textbf{18.43} & 22.19 & \textbf{70.01} & 70.80 & 48.85 & 53.15 \\
CLTS + MTC & 62.67 & 67.19 & 78.44 & 81.33 & 7.61 & 12.60 & 61.04 & 65.90 & 16.25 & 19.93 & 67.47 & 70.00 & 46.85 & 51.63 \\
CLTS + CLP & 64.38 & 68.65 & 78.89 & 82.40 & 8.65 & 13.57 & 61.75 & 67.57 & 17.73 & 20.81 & 68.65 & 71.30 & 47.82 & 53.04 \\
MTC + CLP & 64.04 & 68.92 & 79.16 & 82.56 & 8.81 & 13.05 & 62.26 & 67.34 & 16.97 & 21.31 & 68.98 & 71.20 & 48.06 & 53.32 \\
\midrule
MT + CLTS + MTC & 64.13 & 68.55 & 79.51 & 82.63 & 8.26 & 13.30 & 62.24 & 67.01 & 17.69 & 21.29 & 68.38 & 70.74 & 48.54 & 52.80 \\
MT + CLTS + CLP & 64.50 & 68.82 & \textbf{80.43} & 83.22 & 9.05 & \textbf{13.93} & 62.38 & 67.56 & \textbf{18.66} & \textbf{22.01} & 68.76 & \textbf{71.62} & \textbf{48.96} & \textbf{53.74} \\
MT + MTC + CLP & \textbf{64.71} & \textbf{69.52} & 80.26 & \textbf{83.32} & \textbf{9.15} & 13.37 & \textbf{62.82} & \textbf{67.66} & 17.75 & 21.70 & \textbf{69.33} & 71.50 & 48.93 & 53.41 \\
CLTS + MTC + CLP & 63.81 & 69.00 & 79.70 & 82.52 & 8.69 & 13.48 & 62.25 & 67.04 & 17.55 & 20.75 & 68.23 & 70.96 & 47.68 & 52.55 \\
\midrule
MT + CLTS + MTC + CLP & 64.94 & 68.19 & 80.25 & 82.42 & 9.87 & 13.56 & 62.86 & 67.04 & 16.94 & 21.69 & 68.17 & 71.34 & 49.03 & 53.72 \\
\bottomrule
\end{tabular}}
\caption{Qwen-2.5-14B-Instruct}
\label{tab:rq3_object_b}
\end{subtable}

\caption{\label{tab:diff_object}
Downstream task performance (\%) of LLaMA-3-8B-Instruct and Qwen-2.5-14B-Instruct models trained with different instruction objectives, including MT, CLTS, MTC and CLP.}
\vspace{-1em}
\end{table*}

\subsection{Language Combinations}
\label{sec:rq2_lang_family}
In Section~\ref{sec:rq2_pivot}, we investigate the influence of including English in sampling combinations on model performance.
In this section, we extend that analysis by investigating an alternative sampling strategy: whether the selected languages belong to the same language family.
To this end, we construct four language groups. Groups 1 and 2 consist of languages from the same language family, whereas Groups 3 and 4 include languages from diverse families.
The specific language configurations for each group are detailed in Appendix~\ref{app:sec_lang_comb} (Table~\ref{tab:app_lang_comb}).

\begin{figure}[!h]
    \centering
    \includegraphics[width=\linewidth]{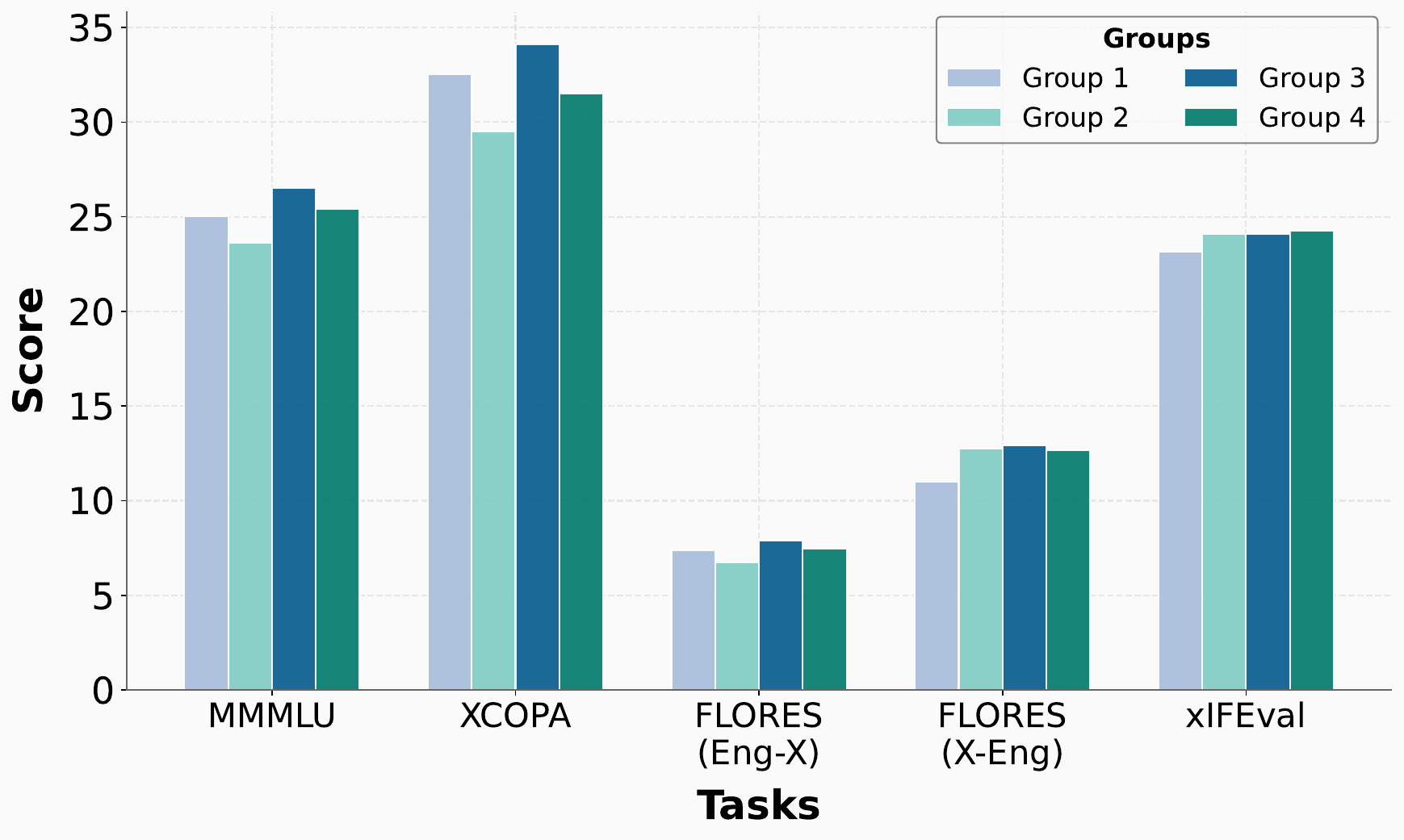}
    \caption{Impact of language family composition on model performance.}
    \label{fig:lang_family}
    \vspace{-1.5em}
\end{figure}

Figure~\ref{fig:lang_family} illustrates the impact of language family composition on model performance.
The results indicate that sampling from cross-family language combinations more effectively leverages the benefits of multi-way parallel corpora, resulting in greater improvements in multilingual task performance.
In contrast, sampling languages from the same family yields only marginal gains. This may be attributed to the structural and lexical similarities among related languages, which can introduce redundancy during training and limit the model's ability to generalize across typologically diverse languages.

%% Figure for different language family

%% figure for different data size
\begin{figure*}[!h]
    \centering
    \includegraphics[width=\linewidth]{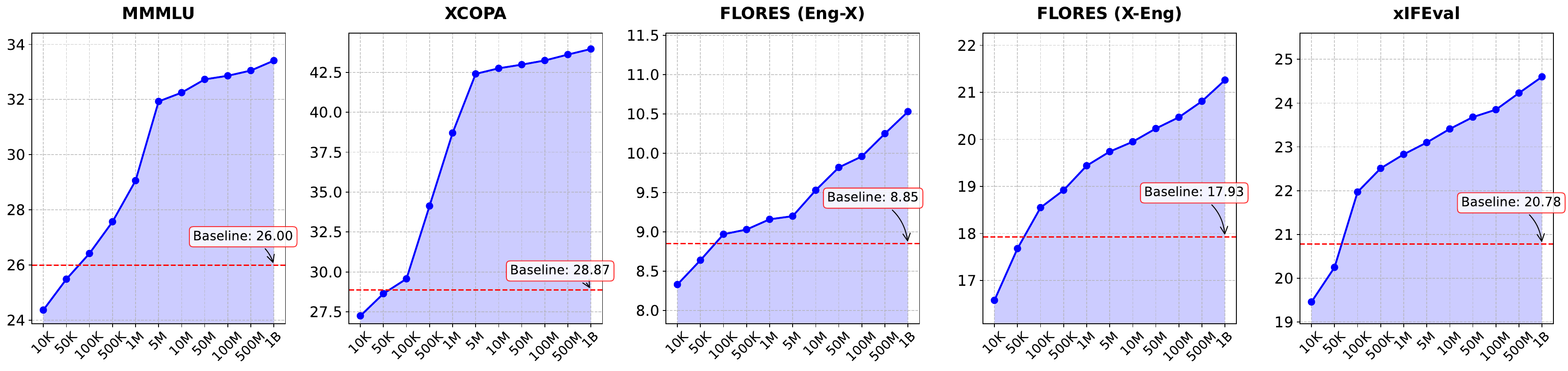}
    \caption{Impact of training data size on model performance across different token amounts (ranging from 10K to 1B) sampled from the \data dataset.}
    \label{fig:data_size}
\end{figure*}

\subsection{Training Data Size}
\label{sec:rq2_data_size}
Figure~\ref{fig:data_size} presents the impact of training data size on model performance.
We conduct experiments by randomly sampling varying amounts of tokens—10K, 50K, 100K, 500K, 1M, 5M, 10M, 50M, 100M, 500M, and 1B—from the constructed \data dataset.
The results demonstrate that model performance is notably constrained when trained on smaller datasets (typically under 100K tokens).
However, as the size of training data increases, performance consistently improves across all evaluated tasks.

For MMMLU and XCOPA, performance exhibits early gains with additional data but plateaus beyond a certain threshold.
This trend likely reflects the nature of these tasks, which emphasize general language understanding and reasoning.
Once the model acquires the necessary core linguistic and world knowledge, the marginal gains from further data diminish.
Interestingly, performance on FLORES and xIFEval continues to improve with increasing data volume.
These tasks, which involve cross-lingual understanding and translation—particularly for low-resource languages or semantically nuanced alignments—appear to benefit more substantially from large-scale data.
This suggests that extensive training data is crucial for enhancing translation quality and evaluation accuracy in such settings.
\section{Instruction Tuning}
\label{sec:rq3}

% %% Prompt Design
\begin{table*}[t]
\resizebox{\textwidth}{!}{
\begin{tabular}{lp{20cm}}
\toprule
\textbf{Task} & \textbf{Prompt} \\
\midrule
Machine Translation (MT) & Translate the following \{src\_lang\_1\}, \{src\_lang\_2\}, ... ,\{src\_lang\_m\} sentence to \{tgt\_lang\_1\}, \{tgt\_lang\_2\}, ..., \{tgt\_lang\_n\}.$\backslash$n \\
& \{src\_lang\_1\} Sentence: \{src\_txt\_1\}.$\backslash$n \{src\_lang\_2\} Sentence: \{src\_txt\_2\}.$\backslash$n ... \{src\_lang\_m\} Sentence: \{src\_txt\_m\}.$\backslash$n \\
& Translation:$\backslash$n \\
& \{tgt\_lang\_1\} Sentence: \{tgt\_txt\_1\}.$\backslash$n \{tgt\_lang\_2\} Sentence: \{tgt\_txt\_2\}.$\backslash$n ... \{tgt\_lang\_n\} Sentence: \{tgt\_txt\_n\}.$\backslash$n \\
\midrule
Cross-Lingual Text Similarity (CLTS) & Given the sentences below in different languages, rate how similar their meanings are on a scale of 0 to 1, where 0 means completely dissimilar and 1 means identical meanings.$\backslash$n \\
& \{lang\_1\} Sentence: \{txt\_1\}.$\backslash$n \{lang\_2\} Sentence: \{txt\_2\}.$\backslash$n ... \{lang\_m\} Sentence: \{txt\_m\}.$\backslash$n \\
& Similarity: \{sim\_score\}.\\
\midrule
Multilingual Text Classification (MTC) & Classify the following sentence in \{lang\_1\}, \{lang\_2\}, ..., \{lang\_m\} into one of the following categories: \{domain\_list\}.$\backslash$n \\
& \{lang\_1\} Sentence: \{txt\_1\}.$\backslash$n \{lang\_2\} Sentence: \{txt\_2\}.$\backslash$n ... \{lang\_m\} Sentence: \{txt\_m\}.$\backslash$n \\
& Categories: \{target\_domain\}.\\
\midrule
Cross-Lingual Paraphrasing (CLP) & Paraphrase the following \{src\_lang\} sentence in \{tgt\_lang\}.$\backslash$n \\
& \{src\_lang\} Sentence: \{src\_txt\}.$\backslash$n \\
& Paraphrasing:$\backslash$n \\
& \{tgt\_lang\} Sentence: \{tgt\_txt\}.\\
\bottomrule
\end{tabular}}
\caption{\label{tab:prompts}
Instruction prompts used for four multilingual tasks: machine translation (MT), cross-lingual text similarity (CLTS), multilingual text classification (MTC), and cross-lingual paraphrasing (CLP). Each prompt is designed to reflect the task's specific objective and structure.
}
\vspace{-1em}
\end{table*}

% In Sections~\ref{sec:rq1} and~\ref{sec:rq2}, we demonstrate that using multi-way data can significantly improve multilingual performance.
In this section, we further investigate whether instruction tuning can also enhance multilingual performance effectively.
Specifically, we address the following key questions\footnote{Additionally, we explore the cross-lingual alignment of representations in instruction tuned models (Appendix~\ref{app:instruct_alignment}.)}:
(1) Which of the different instruction fine-tuning objectives, built on multi-way parallel data, is most effective?
(2) Do models trained with multi-way parallel data exhibit better generalization across domains?

\subsection{Instruction Tuning Objectives}
\label{sec:diff_object}
Using our constructed multi-way parallel data (\data), we define four instruction tasks: machine translation (MT), cross-lingual text similarity (CLTS), multilingual text classification (MTC), and cross-language paraphrasing (CLP).
Table~\ref{tab:diff_object} reports their impact on downstream benchmarks.
Table~\ref{tab:prompts} summarizes the prompt templates and output formats for each task.

We observe that the improvements in MT are the largest and most stable in both high-resource and low-resource languages.
This can primarily be attributed to the fact that, as a token-level supervised generation task, translation strengthens cross-lingual syntactic and semantic consistency, making it a particularly robust task for broad multilingual generalization.
In contrast, CLTS and MTC show smaller drops across different tasks, which we attribute to the coarser granularity of similarity judgments that may not provide fine-grained alignment signals.
Moreover, discrete class labels lack the expressive power to capture subtle semantic distinctions.
Although CLP shows similar advantages to MT in generation tasks, its advantages are narrower in scope and cannot be widely generalized.

Interestingly, the combination of tasks did not significantly improve performance, which may be due to several reasons:
First, the objectives of MT and CLP differ; while MT emphasizes accuracy, CLP focuses more on the diversity of expression, which may make it difficult for the model to balance these two goals during training, thereby affecting performance.
Second, interference between multiple tasks may arise, making it challenging for the model to focus on optimizing the optimal goal of each task, particularly when the task objectives are too similar, amplifying the interference effect.
Furthermore, there is overlap in task design between MT and CLP, which may cause the model to encounter redundant training signals when processing both tasks, preventing it from fully utilizing the unique value of each task.
Finally, joint training of multiple tasks increases the complexity of model training, potentially leading to unstable gradient updates and affecting the model's convergence.

\subsection{Cross-Domain Generalization}
\label{sec:cross_domain}
\begin{figure}[!t]
    \centering
    \vspace{-0.5em}
    \includegraphics[width=\linewidth]{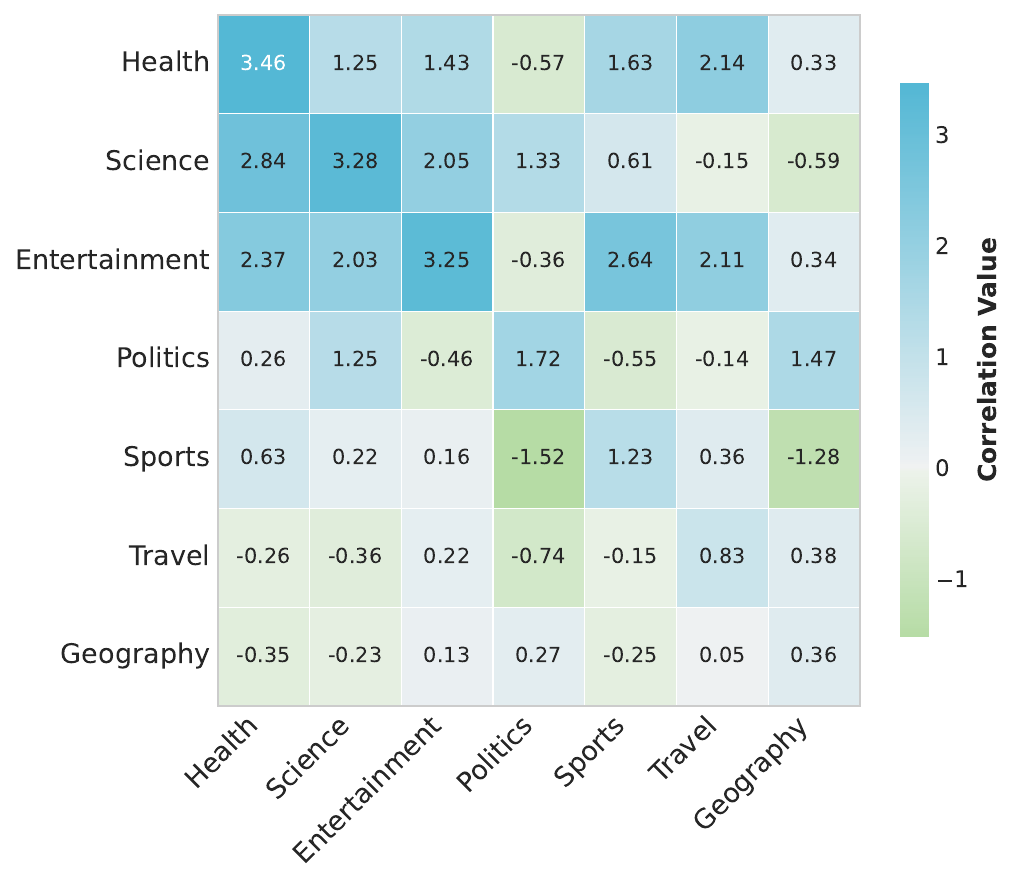}
    \caption{\label{fig:cross_domain}
    Cross-domain generalization performance of instruction-tuned models using multi-way parallel data. Models trained on one domain are evaluated on all other domains from the SIB-200 benchmark.}
    \vspace{-1em}
\end{figure}

To evaluate cross-domain transfer~\cite{lai-etal-2022-improving-domain,liu-etal-2023-compositional}, we extract domain-labeled subsets from \data\ according to the taxonomy of the SIB-200 benchmark.
Instruction tuning is then performed on each domain using the MTC objective, with the resulting models evaluated across all other domains within SIB-200.
Figure~\ref{fig:cross_domain} illustrates the transfer performance.

%% cross domain figure

Overall, instruction tuning with multi-way parallel data significantly improves domain transfer.
The rich cross-lingual and cross-domain signals allow the model to learn domain-invariant features, enhancing robustness when confronted with novel topics and linguistic contexts.
However, transfer performance remains limited in domains such as politics, sports, travel, and geography. We hypothesize that the high topical diversity, coupled with the relative sparsity of domain-specific examples in the training data, hinders the model's ability to capture specialized patterns.
Overcoming these limitations may require more balanced domain coverage or targeted data augmentation strategies.
\section{Conclusion}
\label{sec:conclusion}
In this paper, we construct a large-scale, high-quality multi-way parallel dataset covering 113 languages, with a maximum parallel degree of 50.
This dataset provides a strong foundation for investigating the multilingual adaptation of LLMs.
Using this dataset, we systematically explore best practices for adapting LLMs to multilingual tasks via multi-way parallel data.
Our experiments reveal that multi-way data offers substantial advantages for both continued pretraining and instruction tuning, resulting in improved cross-lingual and cross-domain generalization.
We further analyze key factors influencing model performance, including the degree of parallelism, the language combination strategies, and instruction training objectives.

\section*{Limitations}
This work has the following limitations:
\textbf{(i)} Due to limited computational resources, we employed parameter-efficient fine-tuning (PEFT) methods (LoRA) instead of full-parameter fine-tuning. 
While recent studies have demonstrated that LoRA achieves performance comparable to full fine-tuning across various tasks, our conclusions may still benefit from validation under full fine-tuning or alternative PEFT methods such as adapters or prefix tuning.
\textbf{(ii)} Although our constructed dataset surpasses existing multi-way parallel corpora in both language coverage and maximum parallel degree, its overall size remains modest compared to large-scale unaligned multilingual datasets.
To fully unlock the potential of multi-way parallel data for LLM adaptation, future work will focus on scaling up the dataset to further enhance multilingual performance.
\section*{Acknowledgments}
This work was supported by the National Natural Science Foundation of China (Grants No. 62236011 and No. T2341003), and by a grant from the Guoqiang Institute, Tsinghua University. Additionally, support was received from the European Research Council (ERC) under the European Union’s Horizon Europe Research and Innovation Programme (Grant Agreement No. 101113091), as well as from the German Research Foundation (DFG; Grant FR 2829/7-1).

\bibliography{acl}

\clearpage
\appendix

\section{Statistics of the Constructed \data}
\label{app:dataset}
In Section~\ref{sec:exp}, we introduce \data dataset and present the distribution of sentence counts across languages, along with the overall degree of parallelism (Figure~\ref{fig:parallelism_sta}) and translation quality compared with other multi-way corpora (Figure~\ref{fig:qe_sta}).
To provide a more comprehensive overview, we further analyze domain coverage, fine-grained variations in parallelism, and the distribution of bitexts with respect to both quantity and quality.

\paragraph{Domain Coverage.}
The \data dataset is a multi-domain, multi-way parallel corpus encompassing 352 domains, with domain labels derived from TED Talks.
Table~\ref{tab:app_domain_sta} presents statistics on the number of talks per domain.
We observe that the domains of global issues, education, technology, art, and business constitute the top five in terms of talk count.
This statistical overview offers valuable insights into the dataset’s structure and facilitates a deeper understanding of its composition.
Moreover, the domain labels enhance the dataset's suitability for cross-lingual text classification tasks, positioning \data as a robust benchmark for multilingual text classification.

%%%% Parallelism Sentence
\begin{figure}[h]
    \centering
    \includegraphics[width=\linewidth]{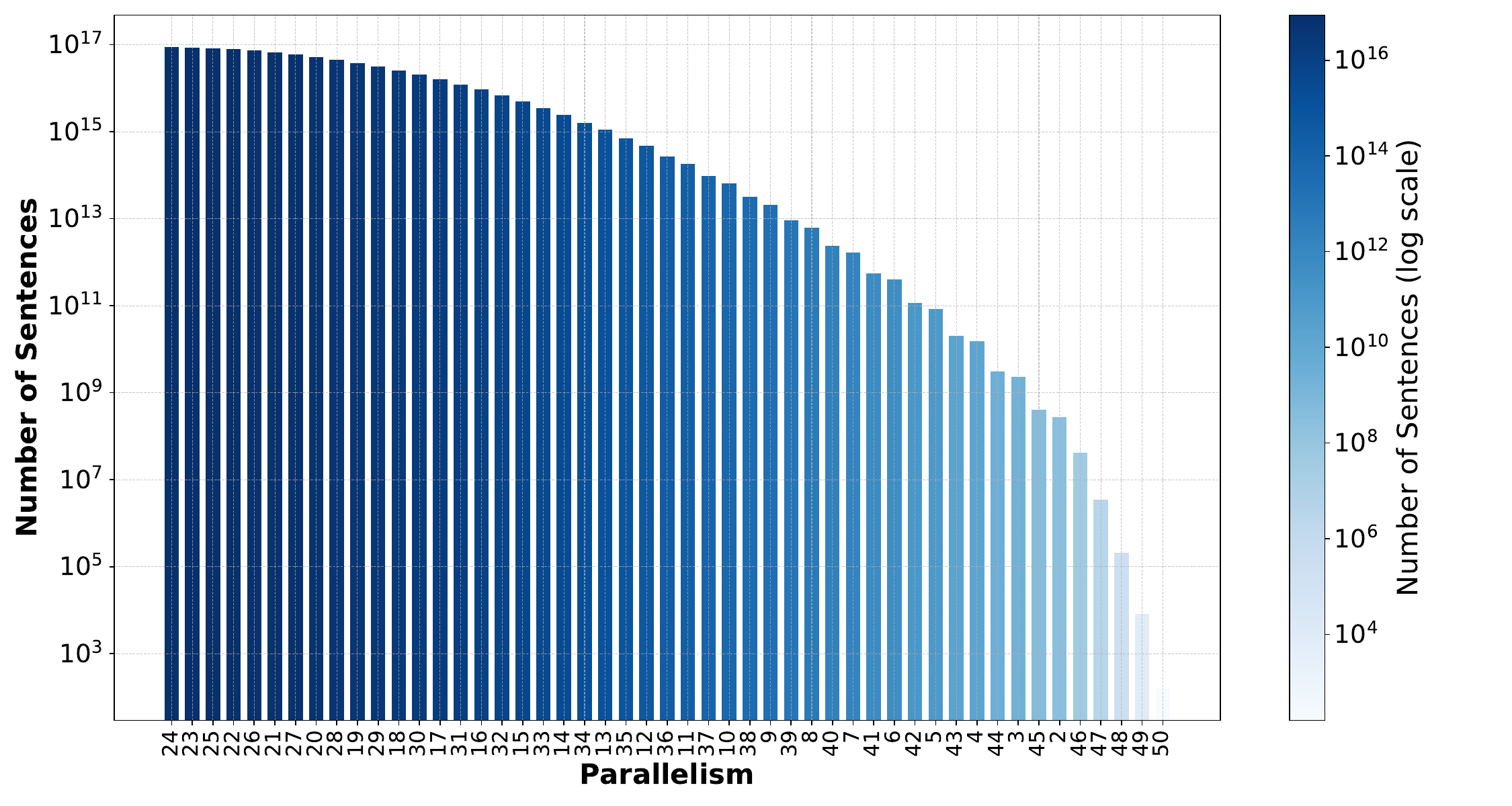}
    \caption{Fine-grained parallelism in \data dataset by tuple size.}
    \label{fig:app_parallel_sents}
\end{figure}

\paragraph{Fine-grained Parallelism.}
In Figure~\ref{fig:parallelism_sta}, we present an approximate count (ratio) of parallelism across different languages.
For a more intuitive understanding of fine-grained parallelism in the \data dataset, Figure~\ref{fig:app_parallel_sents} shows the tuple size corresponding to parallelism.
Note that we count tuples rather than individual sentences.
For instance, in a combination of six languages—English, French, Spanish, Russian, Arabic, and Chinese—a large tuple (en, fr, es, ru, ar, zh) encompasses all possible language combinations.
The total number of such combinations is given by $C_6^1 + C_6^2 + C_6^3 + C_6^4 + C_6^5 + C_6^6 = 6 + 15 + 20 + 15 + 6 + 1 = 63$
In contrast, the corresponding sentence count for this tuple is $6\times1 + 15\times2+ 20\times3 + 15\times4 + 6\times5 + 1\times6 = 192$ sentences.
This tuple-based counting method offers a more precise analysis of parallelism in the dataset.

\paragraph{Quantity and Quality of Bitext.}
In Figure~\ref{fig:qe_sta}, we compare the translation quality of \data with existing multi-way parallel datasets, demonstrating the overall effectiveness of \data translations.
To offer a more comprehensive and intuitive view of translation quality, Table~\ref{tab:app_lang_pair_sta_1},~\ref{tab:app_lang_pair_sta_2},~\ref{tab:app_lang_pair_sta_3},~\ref{tab:app_lang_pair_sta_4},~\ref{tab:app_lang_pair_sta_5}, and~\ref{tab:app_lang_pair_sta_6} report COMET-QE score for all 4,765 language pairs included in the dataset.
This analysis highlights that, despite being a multi-way parallel corpus, \data can be readily decomposed into bilingual sentence pairs, making it suitable for training machine translation models or fine-tuning LLMs on specific language pairs.
By providing both the number of bilingual pairs and their associated translation quality, we aim to support researchers in selecting suitable data for their specific translation tasks and in optimizing performance on targeted language pairs.

\section{Details of Experimental Setup.}
\label{app:exp}

%%%%% Table: parameters for training
\begin{table}[!htbp]
\centering
\begin{tabular}{lclc}
\toprule
\multicolumn{2}{c}{\textbf{LoRA}} & \multicolumn{2}{c}{\textbf{Training}} \\
\cmidrule(lr){1-2}\cmidrule(lr){3-4}
rank          & 8        & batch size      & 8          \\
alpha         & 32       & learning rate   & 1e-04   \\
dropout       & 0.1      & lr schedule     & cosine     \\
target        & all      & warmup ratio    & 0.1        \\
\bottomrule
\end{tabular}
\caption{\label{tab:app_parameters}
Hyper-parameters for continued pretraining and instruction tuning using LLaMA-Factory.
}
\end{table}

\paragraph{Training and Inference Setup.}
Due to computational resource constraints, we adopt LoRA~\cite{hu2022lora} for continued pretraining and instruction fine-tuning of LLMs.
All experiments are conducted using the LLaMA-Factory platform\footnote{\url{https://github.com/hiyouga/LLaMA-Factory}}~\cite{zheng-etal-2024-llamafactory}, with training hyperparameters detailed in Table~\ref{tab:app_parameters}.
Each experiment is run on 8 NVIDIA A100 (80GB) GPUs.
For inference, we utilize the vLLM toolkit\footnote{\url{https://github.com/vllm-project/vllm}}, and the prompt templates used for each benchmark are partially sourced from PromptSource\footnote{\url{https://github.com/bigscience-workshop/promptsource}} as well as the respective original papers.

\paragraph{Evaluation Benchmarks.}
We evaluate the trained model across a diverse set of tasks to comprehensively assess its capabilities in natural language understanding, commonsense reasoning, text generation, instruction following, and text classification.
These tasks span multiple languages and domains, enabling us to measure both general and multilingual performance.
Below, we outline the benchmarks used for each evaluation category:
\begin{itemize}
    \item \textbf{Natural Language Understanding:} We use the MMMLU benchmark, a multilingual extension of the widely adopted MMLU dataset~\cite{hendrycks2020measuring}, designed for evaluating the multitask language understanding abilities of large language models.
    MMMLU covers 14 languages and includes questions from a wide range of domains.
    We report accuracy across all tasks to measure performance.
    \item \textbf{Commonsense Reasoning:} We evaluate the model using the XCOPA dataset~\cite{ponti-etal-2020-xcopa}. XCOPA tests a model’s ability to perform causal commonsense reasoning in multiple languages. The task involves selecting the most plausible cause or effect of a given premise from two alternatives, thus requiring both language understanding and reasoning skills.
    \item \textbf{Text Generation:} We assess multilingual text generation performance using two benchmarks. First, FLORES-101~\cite{goyal-etal-2022-flores} is used to evaluate the model’s general generation quality across a broad set of high- and low-resource languages.
    Second, FLORES-200~\cite{costa2022no} is employed to test zero-shot cross-lingual transfer capabilities—specifically, the model’s ability to generate high-quality outputs in languages it was not directly trained on.
    \item \textbf{Instruction Following:} We use the multilingual variant of the IFEval benchmark~\cite{zhou2023instruction}, implemented by the Benchmax framework~\cite{huang2025benchmax}, to evaluate the model’s ability to follow human instructions across diverse languages and task types. This benchmark focuses on the alignment between user instructions and model responses, which is critical for real-world applications of instruction-tuned models.
    \item \textbf{Text Classification:} For evaluating domain-robust classification performance, we adopt the SIB-200 benchmark~\cite{adelani-etal-2024-sib}, which contains text classification tasks across 200 languages and multiple domains. This benchmark is particularly suited for testing the generalization and robustness of instruction-tuned models in a multilingual setting.
\end{itemize}

%%% Table: non english translation performance
\begin{table*}[t]
\centering
\resizebox{\textwidth}{!}{
\begin{tabular}{l|cccccccccc}
\toprule
          & \multicolumn{2}{c}{sk-ta} & \multicolumn{2}{c}{kk-ko} & \multicolumn{2}{c}{eo-sw} & \multicolumn{2}{c}{ja-zh} & \multicolumn{2}{c}{nl-tr} \\
\cmidrule(lr){2-3}\cmidrule(lr){4-5}\cmidrule(lr){6-7}\cmidrule(lr){8-9}\cmidrule(lr){10-11}
          & BLEU        & COMET       & BLEU        & COMET       & BLEU        & COMET       & BLEU        & COMET       & BLEU        & COMET       \\
\midrule
Baseline  & 3.14        & 51.24       & 3.11        & 50.53       & 2.93        & 48.56       & 5.02        & 54.71       & 3.63        & 52.35       \\
Unaligned & 4.16        & 52.75       & 3.82        & 51.97       & 3.46        & 49.73       & 6.39        & 55.84       & 4.75        & 52.27       \\
Multi-Way & \textbf{6.31}        & \textbf{53.25}       & \textbf{5.38}        & \textbf{52.66}       & \textbf{5.14}        & \textbf{51.35}       & \textbf{8.06}        & \textbf{56.27}       & \textbf{6.82}        & \textbf{53.50}       \\
\bottomrule
\end{tabular}}
\caption{
\label{tab:app_non_eng}
Translation performance of models trained on aligned, unaligned, and multi-way data setups for five non-English-centric language pairs.
}
\end{table*}

\section{Non-English-Centric Translation Performance}
\label{app:non_eng}
In addition to the primary experiments on English–X translation directions presented in Section~\ref{sec:rq1_downstream}, we also investigate the performance of our models on non-English-centric translation pairs.
This analysis is motivated by the potential of multilingual models to perform well in settings where neither language involved is English.
To this end, we randomly select five language pairs (sk-ta, kk-ko, eo-sw, ja-zh, and nl-tr) that do not involve English as either the source or target language.
We then evaluate the translation performance of the pretrained models under three different data configurations: Baseline, Unaligned, and Multi-way, as described in Section~\ref{sec:rq1_downstream}.
The results, including BLEU and COMET scores, are presented in Table~\ref{tab:app_non_eng}.

We observe that:
(1) Models trained on aligned data consistently outperform those trained on unaligned data across all translation pairs, indicating that aligned corpora are essential for achieving high translation quality, regardless of whether English is involved.
(2) Models trained on unaligned data show a significant improvement over the baseline, suggesting that even in the absence of high-quality aligned corpora, leveraging unaligned multilingual data can still enhance translation performance.
(3) The multi-way approach consistently yields superior results compared to both the baseline and unaligned models. This highlights the advantages of multi-way training for multilingual translation tasks, as it not only improves performance on English-centric tasks but also significantly boosts translation quality for non-English-centric language pairs.

% svcca instruction alignment
\begin{figure*}[h]
    \centering
    \includegraphics[width=\linewidth]{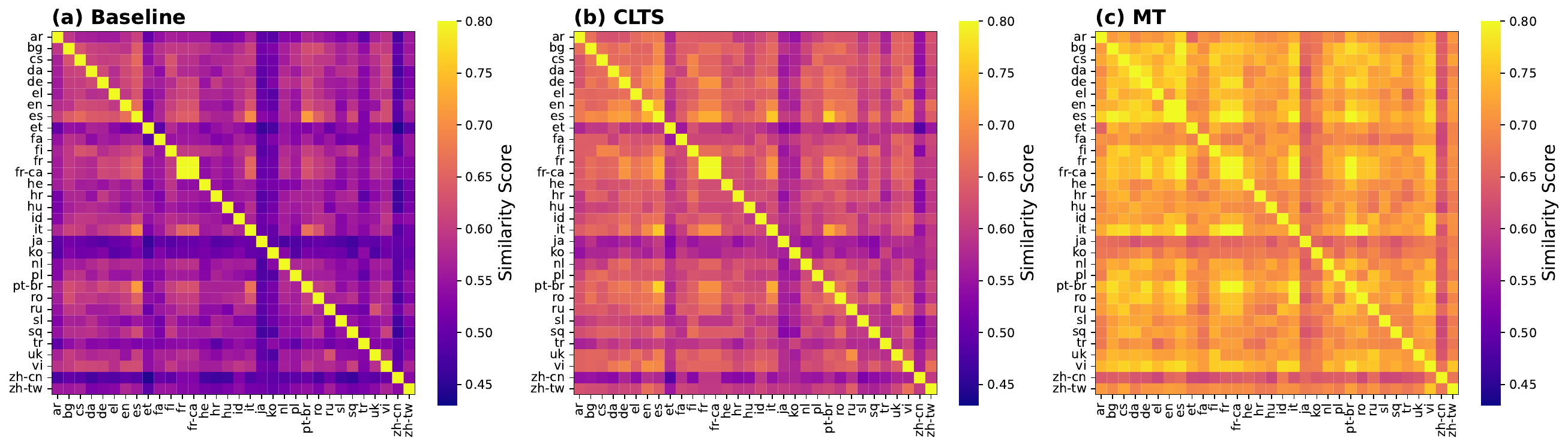}
    \caption{Representation alignment with instruction tuning across different training objectives.}
    \label{fig:svcca_instruct_align}
\end{figure*}

%%%% language combination configuration table
\begin{table*}[!thbp]
\resizebox{\textwidth}{!}{
\begin{tabular}{lll}
\toprule
\multicolumn{3}{c}{\textbf{English Pivoting Combinations}}                                         \\
\midrule
\textbf{Group}   & \multicolumn{2}{l}{\textbf{Language List}}                                               \\
\midrule
group 1 & \multicolumn{2}{l}{en,vi,ar,bg,de,es,fr,he,it,ja,ko}                            \\
group 2 & \multicolumn{2}{l}{en,nl,pl,pt-br,zh-cn,ar,bg,de,es,fr,he}                      \\
group 3 & \multicolumn{2}{l}{en,el,vi,ar,bg,de,es,fr,he,it,ja}                            \\
group 4 & \multicolumn{2}{l}{en,el,ar,bg,de,es,fr,he,it,ja,ko}                            \\
group 5 & \multicolumn{2}{l}{en,fa,hu,ar,bg,de,es,fr,he,it,ja}                            \\
\midrule
\multicolumn{3}{c}{\textbf{Language Family Combinations}}                                          \\
\midrule
\textbf{Group}   & \textbf{Language List}     & \textbf{Language Family}                                             \\
\midrule
group 1 & bg,pl,ru,sr,uk    & Slavic                                                      \\
group 2 & el,it,kmr,sq,tr   & Indo-European                                               \\
group 3 & en,ko,my,sq,zh-tw & Indo-European (en, sq), Sino-Tibetan (zh-tw), Koreanic (ko) \\
group 4 & fr,hu,hy,lv,vi    & Indo-European (fr, hy), Uralic (hu, lv), Austroasiatic (vi) \\
\bottomrule
\end{tabular}}
\caption{\label{tab:app_lang_comb}
Language configurations for different sampling groups.
}
\end{table*}

\section{Configuration of Different Language Combinations}
\label{app:sec_lang_comb}
As discussed in Section~\ref{sec:rq2_lang_family}, we define four distinct language groups for our analysis. The specific configurations of these groups are presented in Table~\ref{tab:app_lang_comb}.

\section{Representation Alignment with Instruction Tuning}
\label{app:instruct_alignment}
In this section, we explore how each tuning objective reshapes the model’s internal multilingual embeddings, using the same alignment metrics as in Section~\ref{sec:rq1_alignment}.
Figure~\ref{fig:svcca_instruct_align} shows that MT-tuned models achieve the highest SVCCA score, indicating tighter alignment among semantically equivalent sentences across languages.
In contrast, CLTS yields minimal alignment gains despite its similarity focus, likely because binary similarity labels lack the contextual richness of translation pairs.

%%% Domain Statistic Table
\begin{table*}[htbp]
\resizebox{\textwidth}{!}{
\begin{tabular}{ll|ll|ll|ll|ll}
\toprule
\textbf{Topic}   & \textbf{\#Talks} & \textbf{Topic}   & \textbf{\#Talks} & \textbf{Topic}   & \textbf{\#Talks} & \textbf{Topic}   & \textbf{\#Talks} & \textbf{Topic}   & \textbf{\#Talks} \\
\midrule
global issues    & 10334   & race & 154     & coronavirus        & 70      & inclusion & 36      & resources  & 18      \\
education        & 9126    & writing & 154     & sex  & 69      & solar system & 35      & paleontology  & 17      \\
technology       & 6323    & physics & 153     & pandemic   & 69      & sight  & 35      & dinosaurs  & 17      \\
art   & 5858    & gender  & 144     & product design     & 68      & Audacious Project    & 35      & television & 17      \\
business & 5694    & exploration & 143     & algorithm  & 68      & industrial design    & 34      & exercise   & 17      \\
Life  & 5623    & neuroscience   & 142     & aging   & 67      & sound  & 34      & blockchain & 17      \\
health   & 5561    & family  & 140     & empathy & 67      & behavioral economics & 34      & fungi  & 16      \\
\bottomrule
\end{tabular}}
\caption{\label{tab:app_domain_sta}
Statistics on the number of talks per domain in the \data dataset.
}
\end{table*}

%%%% language statistics
\begin{table*}[!thbp]
\resizebox{\textwidth}{!}{
\begin{tabular}{lrr|lrr|lrr|lrr|lrr|lrr|lrr}
\toprule
\textbf{Lang Pair}    & \textbf{\#Sents} & \textbf{COMET-QE} & \textbf{Lang Pair}    & \textbf{\#Sents} & \textbf{COMET-QE} & \textbf{Lang Pair}    & \textbf{\#Sents} & \textbf{COMET-QE} & \textbf{Lang Pair}    & \textbf{\#Sents} & \textbf{COMET-QE} & \textbf{Lang Pair}    & \textbf{\#Sents} & \textbf{COMET-QE} & \textbf{Lang Pair}    & \textbf{\#Sents} & \textbf{COMET-QE} & \textbf{Lang Pair}    & \textbf{\#Sents} & \textbf{COMET-QE} \\
\midrule
ar\_en       & 1242114 & 74.58   & ko\_nl    & 603056  & 61.26   & en\_id    & 414970  & 78.6    & bg\_cs    & 290870  & 72.28   & fa\_sq     & 144330  & 62.35   & hi\_ja     & 101055  & 64.02   & ja\_lv       & 77169   & 59.48   \\
en\_es       & 1171895 & 78.54   & nl\_ru    & 602559  & 67.92   & ar\_uk    & 414459  & 63.47   & de\_uk    & 290812  & 70.01   & pt-br\_sq  & 143939  & 65.7    & ca\_fa     & 100797  & 65.67   & ckb\_ru      & 77119   & 66.2    \\
ar\_es       & 1076362 & 67.49   & fa\_fr    & 601008  & 66.3    & el\_nl    & 411252  & 71.5    & cs\_ro    & 289965  & 66.43   & pl\_sq     & 143091  & 58.49   & lt\_th     & 100559  & 64.29   & hy\_it       & 77029   & 72.39   \\
ar\_zh-cn    & 994593  & 56.73   & hu\_ru    & 599111  & 66.95   & el\_ro    & 410986  & 72.64   & ro\_th    & 289049  & 68.12   & sk\_th     & 142947  & 67.93   & kmr\_vi    & 99796   & 57.87   & id\_kmr      & 76830   & 68.53   \\
en\_zh-cn    & 973667  & 68.84   & fa\_zh-cn & 597792  & 58.76   & hr\_zh-tw & 409256  & 65.99   & pt\_tr    & 288876  & 65      & da\_he     & 142920  & 68.94   & my\_ro     & 99349   & 59.51   & lv\_pt-br    & 76810   & 64.23   \\
en\_fr       & 942555  & 78.07   & en\_pl    & 597234  & 71.96   & hr\_ko    & 406822  & 64.74   & id\_nl    & 288406  & 70.38   & es\_lt     & 142342  & 66.23   & da\_uk     & 99161   & 71.01   & hu\_lv       & 76732   & 61.74   \\
\bottomrule
\end{tabular}}
\caption{\label{tab:app_lang_pair_sta_1}
COMET-QE score and the count of sentences for all 4,765 language pairs in \data (part I).
}
\end{table*}

\begin{table*}[!thbp]
\resizebox{\textwidth}{!}{
\begin{tabular}{lrr|lrr|lrr|lrr|lrr|lrr|lrr}
\toprule
\textbf{Lang Pair}    & \textbf{\#Sents} & \textbf{COMET-QE} & \textbf{Lang Pair}    & \textbf{\#Sents} & \textbf{COMET-QE} & \textbf{Lang Pair}    & \textbf{\#Sents} & \textbf{COMET-QE} & \textbf{Lang Pair}    & \textbf{\#Sents} & \textbf{COMET-QE} & \textbf{Lang Pair}    & \textbf{\#Sents} & \textbf{COMET-QE} & \textbf{Lang Pair}    & \textbf{\#Sents} & \textbf{COMET-QE} & \textbf{Lang Pair}    & \textbf{\#Sents} & \textbf{COMET-QE} \\
\midrule
ro\_sl    & 62309 & 65.04 & nb\_th    & 41483 & 68.87 & id\_mr    & 28855 & 62.92 & bn\_vi    & 23238 & 66.1  & en\_sw    & 20023 & 77.74 & mn\_sk    & 17389 & 57.28 & sk\_ta     & 13162 & 62.77 \\
fa\_gl    & 62195 & 66.72 & hy\_sq    & 41172 & 66.65 & de\_mn    & 28786 & 55.66 & az\_ru    & 23136 & 58.27 & it\_kk    & 19975 & 64.07 & hi\_lv    & 17362 & 70.18 & gu\_uk     & 13126 & 67.71 \\
hi\_th    & 61766 & 63.11 & mr\_zh-tw & 41051 & 59.5  & hr\_mr    & 28741 & 61.72 & bs\_tr    & 23103 & 64.35 & kk\_vi    & 19960 & 57.71 & fi\_my    & 17350 & 62.66 & fr-ca\_hi  & 13071 & 67.4  \\
fr-ca\_hu & 61740 & 67.53 & ca\_pt    & 40944 & 67.44 & ar\_ta    & 28701 & 59.23 & fr\_zh    & 23059 & 55.3  & kk\_ru    & 19953 & 62.81 & lt\_mn    & 17323 & 50.67 & fa\_uz     & 12972 & 58.59 \\
nb\_tr    & 61721 & 64.88 & ko\_mr    & 40861 & 57.46 & ru\_ur    & 28670 & 66.47 & es\_zh    & 23042 & 55.89 & eo\_vi    & 19940 & 62.58 & bn\_el    & 17306 & 66.77 & da\_mn     & 12894 & 57.22 \\
gl\_zh-cn & 61551 & 61.4  & mr\_ru    & 40852 & 61.09 & bg\_mr    & 28482 & 59.18 & az\_it    & 23039 & 62.1  & be\_fa    & 19909 & 61.3  & hi\_mk    & 17286 & 66.4  & bs\_sv     & 12879 & 71.79 \\
ckb\_ro   & 61414 & 65.67 & ca\_da    & 40750 & 64.95 & mn\_ro    & 28440 & 54.6  & bs\_fa    & 23027 & 68.83 & lv\_mk    & 19841 & 65.22 & kk\_pl    & 17265 & 64.42 & he\_uz     & 12862 & 56.45 \\
\bottomrule
\end{tabular}}
\caption{\label{tab:app_lang_pair_sta_2}
COMET-QE score and the count of sentences for all 4,765 language pairs in \data (part II).
}
\end{table*}

\begin{table*}[!thbp]
\resizebox{\textwidth}{!}{
\begin{tabular}{lrr|lrr|lrr|lrr|lrr|lrr|lrr}
\toprule
\textbf{Lang Pair}    & \textbf{\#Sents} & \textbf{COMET-QE} & \textbf{Lang Pair}    & \textbf{\#Sents} & \textbf{COMET-QE} & \textbf{Lang Pair}    & \textbf{\#Sents} & \textbf{COMET-QE} & \textbf{Lang Pair}    & \textbf{\#Sents} & \textbf{COMET-QE} & \textbf{Lang Pair}    & \textbf{\#Sents} & \textbf{COMET-QE} & \textbf{Lang Pair}    & \textbf{\#Sents} & \textbf{COMET-QE} & \textbf{Lang Pair}    & \textbf{\#Sents} & \textbf{COMET-QE} \\
\midrule
gl\_nb     & 9763 & 69.76 & it\_ne     & 7276 & 68.96 & ne\_uk     & 5902 & 67.79 & ca\_sw     & 4657 & 64.38 & kn\_zh-tw  & 3939 & 57.23 & en\_kn     & 2984 & 69.3  & ka\_uz    & 2385 & 57.03 \\
sw\_th     & 9717 & 64.26 & fil\_tr    & 7235 & 55.82 & eo\_lv     & 5847 & 60.91 & ar\_sh     & 4655 & 68.13 & de\_kn     & 3937 & 62.02 & bn\_ckb    & 2975 & 64.63 & fi\_ne    & 2380 & 66.46 \\
lv\_mr     & 9713 & 60.05 & fil\_ja    & 7210 & 57.82 & id\_ml     & 5817 & 63.08 & az\_my     & 4643 & 62.26 & cs\_srp    & 3933 & 68.93 & cs\_si     & 2972 & 63.69 & ja\_tl    & 2370 & 58.68 \\
ckb\_sq    & 9677 & 56.9  & eu\_mk     & 7177 & 59.87 & fil\_id    & 5816 & 61.9  & ckb\_gu    & 4633 & 62.64 & pt\_srp    & 3924 & 71.02 & cs\_kn     & 2971 & 58.32 & de\_tl    & 2361 & 61.46 \\
kk\_pt     & 9660 & 64.35 & te\_th     & 7167 & 60.45 & ne\_th     & 5800 & 62.89 & is\_zh-tw  & 4615 & 58    & kn\_tr     & 3923 & 61.92 & mn\_uz     & 2970 & 54.7  & pt-br\_tl & 2361 & 62.52 \\
fi\_mr     & 9659 & 61.05 & et\_kmr    & 7154 & 59.83 & eu\_pt     & 5798 & 64.97 & sr\_srp    & 4608 & 71.38 & is\_pl     & 3913 & 61.64 & bs\_nb     & 2969 & 70.56 & el\_kn    & 2357 & 58.97 \\
ar\_te     & 9648 & 57.75 & sq\_uz     & 7149 & 49.5  & az\_kmr    & 5783 & 54.73 & hu\_is     & 4598 & 59.21 & bg\_kn     & 3904 & 60.9  & lt\_ml     & 2964 & 57.47 & eu\_mn    & 2354 & 52.06 \\
\bottomrule
\end{tabular}}
\caption{\label{tab:app_lang_pair_sta_3}
COMET-QE score and the count of sentences for all 4,765 language pairs in \data (part III).
}
\end{table*}

\begin{table*}[!thbp]
\resizebox{\textwidth}{!}{
\begin{tabular}{lrr|lrr|lrr|lrr|lrr|lrr|lrr}
\toprule
\textbf{Lang Pair}    & \textbf{\#Sents} & \textbf{COMET-QE} & \textbf{Lang Pair}    & \textbf{\#Sents} & \textbf{COMET-QE} & \textbf{Lang Pair}    & \textbf{\#Sents} & \textbf{COMET-QE} & \textbf{Lang Pair}    & \textbf{\#Sents} & \textbf{COMET-QE} & \textbf{Lang Pair}    & \textbf{\#Sents} & \textbf{COMET-QE} & \textbf{Lang Pair}    & \textbf{\#Sents} & \textbf{COMET-QE} & \textbf{Lang Pair}    & \textbf{\#Sents} & \textbf{COMET-QE} \\
\midrule
fil\_sl    & 1886 & 59.82 & gu\_ta    & 1557 & 66.39 & ja\_pa    & 1304 & 63.13 & hy\_srp   & 1082 & 65.79 & bg\_lo    & 854 & 64.75 & pa\_uk     & 665 & 62.48 & ta\_tg    & 532 & 44.58 \\
km\_ro     & 1881 & 67.57 & ps\_pt    & 1554 & 68.04 & arq\_my   & 1298 & 53.16 & kn\_mn    & 1082 & 57.54 & eu\_lo    & 854 & 58.47 & si\_ur     & 665 & 68.2  & lo\_nl    & 530 & 69.82 \\
kn\_lt     & 1878 & 58.29 & th\_tl    & 1553 & 61.61 & mk\_nn    & 1296 & 70.71 & ml\_my    & 1079 & 59.78 & et\_lo    & 854 & 62.6  & ja\_sc     & 664 & 45.6  & af\_be    & 524 & 59.74 \\
arq\_pl    & 1870 & 46.11 & ca\_srp   & 1552 & 67.46 & id\_ug    & 1296 & 69.21 & tg\_uk    & 1079 & 37.69 & lo\_tr    & 854 & 62.66 & bs\_ne     & 663 & 67.47 & srp\_ur   & 523 & 66.31 \\
bo\_ru     & 1870 & 41.9  & ms\_srp   & 1549 & 66.72 & af\_da    & 1294 & 63.71 & ht\_ko    & 1076 & 46.13 & lo\_th    & 854 & 64.58 & cs\_pa     & 661 & 69.44 & es\_lo    & 523 & 65.34 \\
bo\_de     & 1868 & 41.21 & eu\_mr    & 1547 & 57.91 & ms\_so    & 1290 & 46.86 & pa\_th    & 1073 & 67.43 & ar\_lo    & 854 & 61.37 & mg\_vi     & 661 & 48.29 & lo\_ru    & 523 & 59.54 \\
hi\_nn     & 1865 & 71.95 & pa\_zh-tw & 1542 & 65.66 & ckb\_srp  & 1283 & 64.88 & gl\_nn    & 1073 & 65.89 & da\_lo    & 854 & 68.15 & fr\_mg     & 661 & 45.43 & eo\_kn    & 522 & 65.77 \\
\bottomrule
\end{tabular}}
\caption{\label{tab:app_lang_pair_sta_4}
COMET-QE score and the count of sentences for all 4,765 language pairs in \data (part IV).
}
\end{table*}

\begin{table*}[!thbp]
\resizebox{\textwidth}{!}{
\begin{tabular}{lrr|lrr|lrr|lrr|lrr|lrr|lrr}
\toprule
\textbf{Lang Pair}    & \textbf{\#Sents} & \textbf{COMET-QE} & \textbf{Lang Pair}    & \textbf{\#Sents} & \textbf{COMET-QE} & \textbf{Lang Pair}    & \textbf{\#Sents} & \textbf{COMET-QE} & \textbf{Lang Pair}    & \textbf{\#Sents} & \textbf{COMET-QE} & \textbf{Lang Pair}    & \textbf{\#Sents} & \textbf{COMET-QE} & \textbf{Lang Pair}    & \textbf{\#Sents} & \textbf{COMET-QE} & \textbf{Lang Pair}    & \textbf{\#Sents} & \textbf{COMET-QE} \\
\midrule
en\_mg     & 411 & 57.44 & id\_mt    & 343 & 45.08 & ga\_ru     & 274 & 52.87 & lb\_nl     & 217 & 40.42 & ug\_zh     & 159 & 62.63 & am\_ur     & 128 & 66.7  & km\_sl     & 94 & 62.4  \\
km\_ky     & 411 & 62.79 & mn\_ug    & 343 & 56.64 & kn\_nb     & 272 & 63.34 & hu\_lb     & 217 & 38.76 & tl\_ug     & 159 & 63    & ceb\_uk    & 128 & 48.32 & ckb\_is    & 93 & 61.37 \\
eu\_ky     & 411 & 59.86 & tg\_tl    & 342 & 38.36 & ga\_ro     & 271 & 50.93 & hr\_lb     & 217 & 39.3  & hy\_pa     & 158 & 60.44 & lv\_ps     & 128 & 61.37 & ckb\_fil   & 93 & 53.04 \\
eo\_ky     & 411 & 60.48 & en\_mt    & 340 & 64.14 & km\_sk     & 271 & 70.12 & he\_lb     & 217 & 38.13 & el\_tt     & 157 & 55.9  & ig\_pt     & 127 & 38.76 & fo\_my     & 93 & 50.27 \\
am\_ky     & 411 & 59.28 & et\_mt    & 340 & 42.85 & bn\_ps     & 271 & 64.38 & lb\_pt-br  & 217 & 41.76 & cs\_tt     & 156 & 57.62 & mg\_pt     & 125 & 49.39 & fr\_ig     & 92 & 36    \\
ky\_ta     & 411 & 68.26 & fr-ca\_mt & 340 & 45.25 & ar\_ga     & 270 & 51.87 & lb\_my     & 217 & 49.83 & fr\_tt     & 156 & 55.26 & am\_ms     & 125 & 60.65 & fo\_fr     & 92 & 51.41 \\
am\_km     & 411 & 60.48 & da\_mt    & 340 & 46.18 & ga\_my     & 269 & 53.86 & te\_uz     & 217 & 59.9  & fa\_tt     & 156 & 59.29 & af\_bo     & 122 & 36.83 & fo\_sr     & 92 & 48.98 \\
\bottomrule
\end{tabular}}
\caption{\label{tab:app_lang_pair_sta_5}
COMET-QE score and the count of sentences for all 4,765 language pairs in \data (part V).
}
\end{table*}

\begin{table*}[!thbp]
\resizebox{\textwidth}{!}{
\begin{tabular}{lrr|lrr|lrr|lrr|lrr|lrr|lrr}
\toprule
\textbf{Lang Pair}    & \textbf{\#Sents} & \textbf{COMET-QE} & \textbf{Lang Pair}    & \textbf{\#Sents} & \textbf{COMET-QE} & \textbf{Lang Pair}    & \textbf{\#Sents} & \textbf{COMET-QE} & \textbf{Lang Pair}    & \textbf{\#Sents} & \textbf{COMET-QE} & \textbf{Lang Pair}    & \textbf{\#Sents} & \textbf{COMET-QE} & \textbf{Lang Pair}    & \textbf{\#Sents} & \textbf{COMET-QE} & \textbf{Lang Pair}    & \textbf{\#Sents} & \textbf{COMET-QE} \\
\midrule
bo\_ga     & 66 & 39.42 & as\_uz    & 57 & 56.78 & ast\_eo    & 54 & 56.27 & bi\_fr     & 49 & 38.21 & gl\_la    & 33 & 46.56 & ro\_szl   & 23 & 38.62 & ne\_sw     & 4 & 77.92 \\
ga\_sv     & 66 & 62.57 & as\_fi    & 57 & 68.52 & eo\_oc     & 54 & 37.63 & bi\_fi     & 49 & 39.17 & ga\_lv    & 33 & 54.87 & kmr\_mg   & 22 & 47.65 & gu\_hup    & 3 & 39.39 \\
ga\_sl     & 66 & 53.98 & as\_uk    & 57 & 64.42 & en\_oc     & 54 & 45.12 & bi\_uk     & 49 & 43.11 & ga\_la    & 33 & 46.36 & oc\_ro    & 22 & 30.58 & gu\_mt     & 3 & 56.14 \\
szl\_vi    & 66 & 37.79 & as\_bg    & 57 & 63.52 & oc\_tr     & 54 & 42.44 & bi\_fa     & 49 & 45.28 & ga\_kk    & 33 & 50.93 & lv\_tlh   & 22 & 34.58 & ckb\_hup   & 3 & 38.44 \\
cs\_szl    & 66 & 34.42 & as\_ko    & 57 & 63.2  & ast\_az    & 54 & 53.9  & bg\_bi     & 49 & 33.37 & eo\_ga    & 33 & 53.9  & ka\_tlh   & 22 & 39.07 & hup\_vi    & 3 & 36.74 \\
lt\_szl    & 66 & 37.71 & as\_es    & 57 & 67.51 & az\_oc     & 54 & 41.61 & bi\_ko     & 49 & 48.18 & ga\_tl    & 33 & 48.45 & ig\_ta    & 22 & 38.14 & cs\_hup    & 3 & 29.79 \\
lo\_szl    & 66 & 47.98 & as\_en    & 57 & 77.97 & oc\_th     & 54 & 42.46 & bi\_kk     & 49 & 48.99 & ga\_ta    & 33 & 58.61 & ha\_ro    & 22 & 48.38 & hup\_lv    & 3 & 33.43 \\
\bottomrule
\end{tabular}}
\caption{\label{tab:app_lang_pair_sta_6}
COMET-QE score and the count of sentences for all 4,765 language pairs in \data (part VI).
}
\end{table*}

\end{document}